\RequirePackage{fix-cm}
\documentclass[twocolumn]{svjour3}          
\smartqed  

\journalname{Autonomous Robots}

\usepackage[sort&compress]{natbib}
\usepackage[colorlinks=true,citecolor=blue,linkcolor=blue,filecolor=blue,urlcolor=blue]{hyperref}
\usepackage{graphicx}
\usepackage{amsmath}
\usepackage{url}
\usepackage{comment} 
\usepackage{amssymb}
\usepackage{color}

\DeclareMathOperator*{\argmax}{argmax}

\usepackage{type1cm}

\usepackage{algorithm}
\usepackage{algpseudocode}
\algrenewcommand{\algorithmicreturn}{\State \textbf{return}}

\begin{document}
\title{
Improved and Scalable Online Learning of \\ Spatial Concepts and Language Models with Mapping
\thanks{This work was partially supported by JST CREST grant number JPMJCR15E3, and JSPS KAKENHI grant number JP17J07842, JP16H06561, and JP16K12497.}
}


\titlerunning{Improved and Scalable Online Learning}

\author{Akira Taniguchi \and Yoshinobu Hagiwara \and Tadahiro Taniguchi \and Tetsunari~Inamura}

\institute{Akira Taniguchi \and Yoshinobu Hagiwara \and Tadahiro Taniguchi \at
              Ritsumeikan University, 1-1-1 Noji-Higashi, Kusatsu, Shiga 525-8577, Japan. \\
              \email{\{a.taniguchi, yhagiwara, taniguchi\} @em.ci.ritsumei.ac.jp}           
           \and
           Tetsunari Inamura \at
              the National Institute of Informatics / SOKENDAI (The Graduate University for Advanced Studies), 2-1-2 Hitotsubashi, Chiyoda-ku, Tokyo 101-8430, Japan.
              \email{inamura@nii.ac.jp}           
}

\date{Received: date / Accepted: date}

\maketitle

\begin{abstract}
We propose a novel online learning algorithm, called SpCoSLAM 2.0, for spatial concepts and lexical acquisition with high accuracy and scalability.
Previously, we proposed SpCoSLAM as an online learning algorithm based on unsupervised Bayesian probabilistic model that integrates multimodal place categorization, lexical acquisition, and SLAM.
However, our original algorithm had limited estimation accuracy owing to the influence of the early stages of learning, and increased computational complexity with added training data.
Therefore, we introduce techniques such as fixed-lag rejuvenation to reduce the calculation time while maintaining an accuracy higher than that of the original algorithm.
The results show that, in terms of estimation accuracy, the proposed algorithm exceeds the original algorithm and is comparable to batch learning.
In addition, the calculation time of the proposed algorithm does not depend on the amount of training data and becomes constant for each step of the scalable algorithm.
Our approach will contribute to the realization of long-term spatial language interactions between humans and robots.
\end{abstract}


\begin{figure*}[!tb]
  \begin{center}
    \includegraphics[width=0.85\hsize]{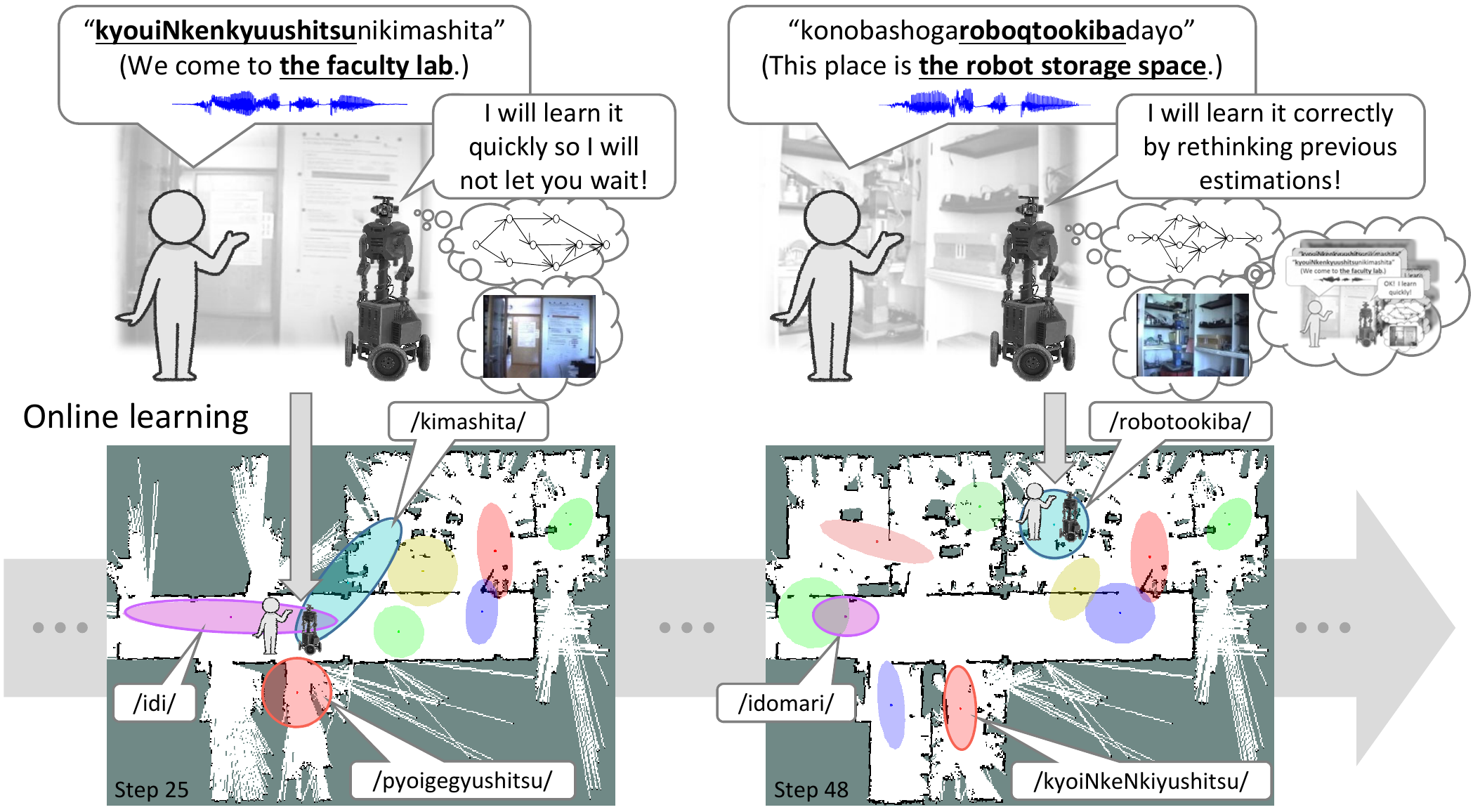}
    \caption{Overview of the scenario for online learning in this study. 
    We assume a scenario in which the user teaches the robot the name of the place using a spoken utterance while moving together in the environment.
    The robot learns spatial concepts, language models, and maps while sequentially correcting mistakes from previous learnings based on its interaction with the user and environment, as shown from the bottom left to the bottom right.}
    \label{fig:overview}
  \end{center}
\end{figure*}

\section{Introduction}
\label{sec:introduction}
Robots operating in various human environments must adaptively and sequentially acquire new categories for places and unknown words related to various places as well as the map of the environment \citep{kostavelis2015semantic}.
It is desirable for robots to acquire place categories and vocabulary autonomously based on their experience because it is difficult to manually design spatial knowledge in advance.
Related research in the fields of semantic mapping and place categorization \citep{pronobis2012large,kostavelis2015semantic,sunderhauf2016place,landsiedel2017review,Rangel2018} has attracted considerable interest in recent years.
However, conventional approaches in most of these studies are limited insofar as the robots cannot learn unknown words and unknown place categories without pre-set vocabulary and categories. 
In addition, the processes for Simultaneous Localization And Mapping (SLAM) \citep{thrun2005probabilistic} and for estimating semantics related to place have been addressed as separated module processes. 
However, in our proposed approach, the robot can automatically and simultaneously perform place categorization and environment mapping, and it can learn unknown words without prior knowledge.
Our previously proposed unsupervised Bayesian probabilistic model integrates multimodal place categorization, lexical acquisition, and SLAM.
In particular, this paper focuses on the problems of estimation accuracy and computational scalability in online learning.

We define a spatial concept as a place category is autonomously learned by the robot based on multimodal perceptual information, which includes names of places, features of scene images, and position distributions. 
Then, we define a position distribution as the spatial extent representing a place in the environment.
Our study regarding the spatial concept formation and the lexical acquisition also constitute constructive approaches to the human developmental process and symbol emergence in cognitive developmental systems \citep{cangelosi2015developmental,taniguchi2018TCDSsurvey}.
Thus, we assume that the robot has not acquired any vocabulary in advance and can recognize only phonemes or syllables.
In addition, the robot does not have prior knowledge of the current environment. 
In this study, a scenario in which the user teaches the robot the name of a place using a spoken utterance while moving together in the environment is studied.
An overview of the scenario for online learning task is shown in Fig.~\ref{fig:overview}. 
The robot and the user move around the environment. 
When they come to a place where the user wishes to teach, the user speaks a sentence regarding the place to the robot.
The robot recognizes the speech, including unknown words, and segments the speech into words. 
Then, the robot obtains the present estimated position, the scene image, and the speech signal at that time, and acquires spatial knowledge regarding the environment, such as the relationship between words and places.

In online learning, also called sequential learning or incremental learning, an increase in scalability without reducing accuracy is especially important but difficult to achieve for mobile robots.
Online learning has the advantage of being performed in real-time. 
This means that it can be used to adapt immediately to new data by sequentially estimating parameters each time.
On the other hand, batch learning takes time to collect large amounts of data and to iterate it for learning.
In the case of online learning, previous knowledge can be used immediately for reasoning and tasks such as language communication.
\citet{ataniguchi_IROS2017} focused on deriving and constructing an appropriate online learning algorithm mathematically based on a theory of machine learning.
In our previous work, we proposed SpCoSLAM as an integrated model of nonparametric Bayesian multimodal categorization, a Bayesian filter-based SLAM, speech recognition, and word segmentation, from the standpoint of unsupervised machine learning.
However, this algorithm \citep{ataniguchi_IROS2017} had inferior accuracy in terms of categorization and word segmentation compared to batch learning, owing to a situation whereby sufficient statistical information could not be used at the early stages of learning.
In addition, speech recognition and unsupervised word segmentation were not completely online, and batch learning was used as an approximation.
Therefore, the computational complexity of the processes of speech recognition and unsupervised word segmentation increased with an increase in training data.
To enable online learning based on long-term human--robot interactions with limited computational resources, 
the following core problems need to be solved: (i) the increase in calculation cost owing to an increase in data, and (ii) the decrease in estimation accuracy when compared with batch learning.
In intelligent robotics, the framework of online learning is regarded as important.
In particular, online learning, which solves the above problem, is required for robots that gain knowledge while moving in the real world.

We here describe improved and scalable algorithms to solve the above-mentioned problems.
The improved algorithm mainly addresses the problems of misrecognition (misclassification) and word segmentation in online learning.
The scalable algorithm mainly addresses the problem of the increase in computation time.
In this study, we introduce the approach of fixed-lag rejuvenation, which is considered particularly effective at solving these problems.
Regarding the problem of online lexical acquisition, the improved and scalable algorithms take two respective approaches to the solution.
The improved algorithm addresses the problem of under-segmentation, whereby the phoneme sequence is insufficiently segmented, by changing the manner by which the language model is updated such that it re-segments the word sequence.
The scalable algorithm performs in a pseudo-online manner by introducing a fixed-lag rejuvenation approach to speech recognition and word segmentation.

One of the advantages to the proposed online learning algorithm is that spatial concepts mistakenly learned by the robot can be corrected sequentially, something that could not be achieved thus far.
Moreover, with the proposed algorithm, the robot can flexibly deal with changes in the environment and the names of places.
The lower part in Fig.~\ref{fig:overview} shows the progress of online learning.
In the lower left of Fig.~\ref{fig:overview}, clustered places and words are incorrectly estimated, as shown by the elongated purple and blue ellipses.
In the lower right of Fig.~\ref{fig:overview}, by contrast, more accurate estimation is achieved by correcting errors as learning progresses.
This is realized by reviewing and rethinking previous estimation results when new data is obtained.

The main contributions of this paper are as follows:
\begin{itemize}
 \item       
We propose an improved and scalable online learning algorithm with several novel techniques such as fixed-lag rejuvenation.
 \item       
The improved online algorithm achieves an accuracy of place categorization and lexical acquisition comparable to batch learning.
 \item       
The scalable online algorithm achieves faster learning compared to original algorithms by reducing the order of computational complexity.
\end{itemize}

\newsavebox{\boxb} \sbox{\boxb}{\ref{sec:exp2}}

The remainder of this paper is organized as follows.
In Section~\ref{sec:RelatedWork}, we discuss related work on the formation of spatial concepts and online learning that is relevant to our study.
In Section~\ref{sec:SpCoSLAM}, we present an overview of the model, along with the formulation and the original online learning algorithm, SpCoSLAM.
In Section~\ref{sec:proposed}, we present our proposed algorithms for improved and scalable online learning.
In Section~\ref{sec:exp}, we discuss the effectiveness of the proposed algorithms in a real environment.
{In Section~{\usebox\boxb}, we evaluate the performance of place categorization and lexical acquisition in various virtual home environments.}
Section~\ref{sec:conclusion} concludes the paper.

\section{Related work}
\label{sec:RelatedWork}

\subsection{Spatial concept formation}
\label{sec:RelatedWork:spatial}
\citet{taguchi2011learning} proposed an unsupervised method for simultaneously categorizing self-positions and phoneme sequences from user speech without any prior language model.
\citet{taniguchi_spcoa,taniguchi2018unsupervised} proposed the nonparametric Bayesian Spatial Concept Acquisition method (SpCoA) using an unsupervised word segmentation method, latticelm \citep{neubig2012bayesian}, and SpCoA++ for highly accurate lexical acquisition as a result of updating the language model.
\citet{gu2016learning} proposed a method to learn relative spatial concepts, i.e., the words related to distance and direction, from the positional relationship between an utterer and objects. 
\citet{isobe2017learning} proposed a learning method to derive the relationship between objects and places using image features obtained by a Convolutional Neural Network (CNN) \citep{krizhevsky2012imagenet}.
\citet{hagiwara2018hierarchical} implemented a hierarchical clustering method for the formation of hierarchical place concepts.
However, none of the above methods can sequentially learn spatial concepts from unknown environments without a map, because they rely on batch-learning algorithms.
Therefore, we developed in previous work an online algorithm, SpCoSLAM \citep{ataniguchi_IROS2017}, that can sequentially learn a map, a lexicon, and spatial concepts to integrate positions, speech signals, and scene images. 
In \citet{ataniguchi_IROS2017}, however, the accuracy was inferior to that of SpCoA.
In this paper, we also compare our proposal to the latest batch learning method, SpCoA++.
Because SpCoA++ is able to achieve nearly correct lexical acquisition, if we can successfully overcome the above problems by appropriately devising the learning algorithm, its accuracy should improve even with online lexical acquisition.

Our approach is relevant to research integrating semantic mapping with natural language processing \citep{walter2013learning,hemachandra2014learning}.
\citet{walter2013learning} developed an algorithm that can learn semantic graphs to integrate semantic representation into metric maps from natural language descriptions of aspects such as labels and spatial relationships. 
\citet{hemachandra2014learning} proposed a mechanism to more effectively ground natural language descriptions by integrating scene appearance observations using camera images and laser data.
In these studies, a word list, place labels, and the number of category types were known in advance.
However, it is challenging to sequentially acquire new words and categories efficiently from a situation in which the lists of words and categories are not provided in advance. Our study includes lexical acquisition for unknown words and formation of new categories from speech signals using spatial information.

\citet{ball2013openratslam} implemented a biologically inspired mapping system, RatSLAM, which is related to pose cells in the hippocampus of a rodent. 
In addition, robots called Lingodroids using RatSLAM could acquire a lexicon related to places through robot-to-robot communication \citep{heathlingodroids2016}.
These studies reported that robots created their own vocabulary. 
\citet{ueda2016particle} proposed a brain-inspired method, namely, a Particle Filter on Episode (PFoE) for agent decision making.
PFoE can estimate the agent's internal state based on previous events recalled at the time.
All previous data is thus accumulated to construct a state space in PFoE.
We believe that PFoE is unsuitable for long-term trials because the state space becomes enormous.
By contrast, our approach forms concepts from episodes using resources more reasonably for calculations, insofar as the state space is reduced through clustering.
Although our proposed method was not originally inspired by biology or brain science, such research is highly suggestive.
SpCoSLAM is an integrated model of self-localization, mapping, concept formation, and lexical acquisition.
From the point of view of the brain, it may be possible to regard SpCoSLAM as a model that imitates some functions of the hippocampus and the cerebral cortex.
If we assume that the training data---i.e., the robot's experiences based on a user's utterances---is the episodic memory, and that spatial concepts are semantic memory, the proposed algorithm can be interpreted as a representation of the process of forming concepts by extracting meaning from short-term episodic memory sequentially.
Such matters are not further discussed in this paper, although they remain important for future research.

\begin{figure*}[!tb]
\begin{center}
  \begin{minipage}{.34\hsize}
    \begin{center}
    \includegraphics[width=0.962\hsize]{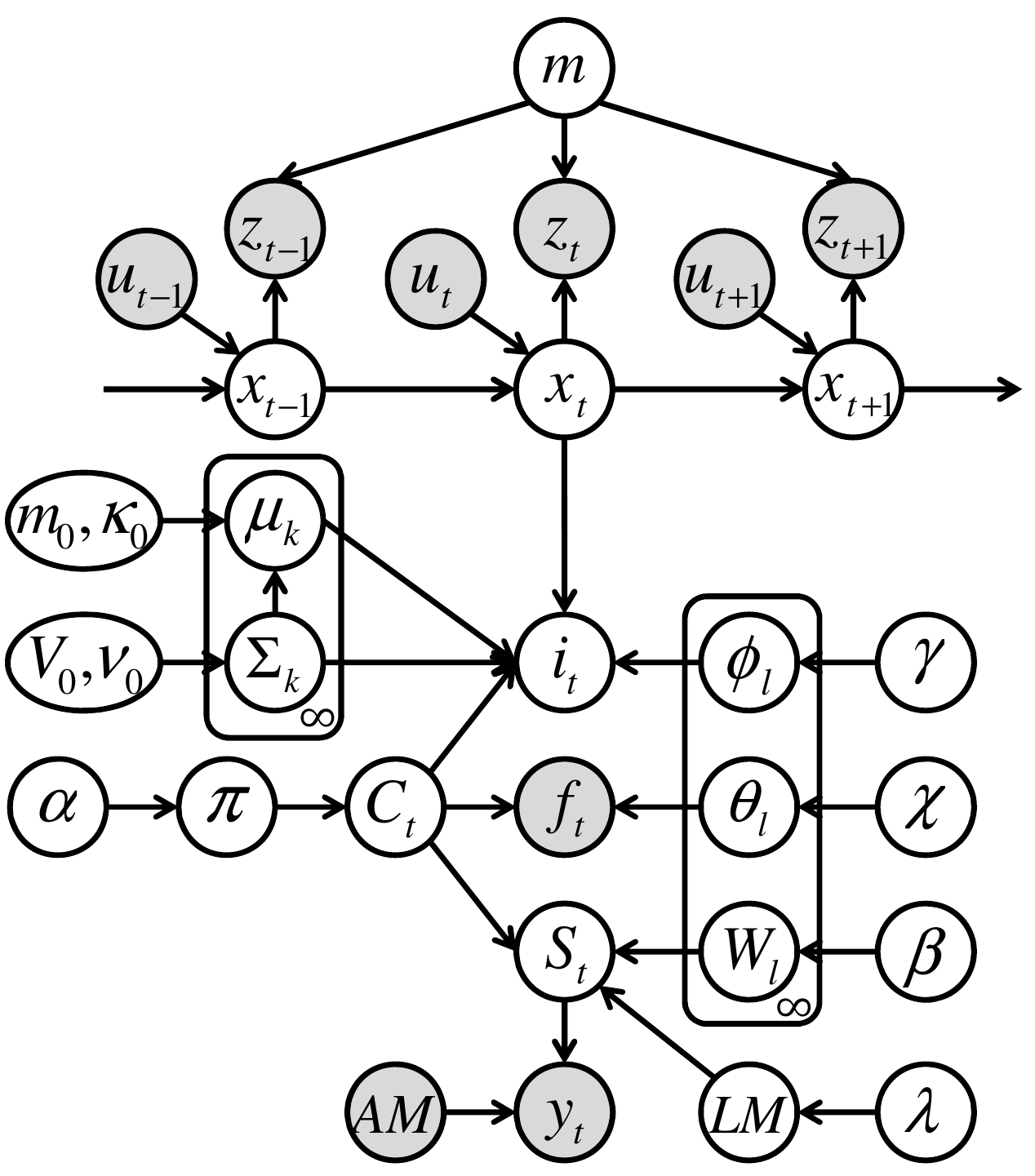}
    \end{center}
  \end{minipage}
  \begin{minipage}{.64\hsize}
\begin{center}
\begin{tabular}{lp{240pt}} \hline 
Symbol & Definition \\ \hline
$m$ & Environmental map \\ 
$x_t$ & Self-position of a robot \\ 
$z_t$ & Depth data \\ 
$u_t$ & Control data \\ 
$f_{t}$ & Image feature \\ 
$y_{t}$ & Speech signal \\ 
$i_{t}$ & Index of position distributions \\ 
$C_{t}$ & Index of spatial concepts \\ 
{$S_t$} & {Word sequence (word segmentation result)} \\ 
{$\mu_{k}$, $\Sigma_{k}$} & {Parameters of Gaussian distribution (position distribution)} \\ 
{$\pi $} & {Parameter of multinomial distribution for index $C_t$ of spatial concepts} \\ 
{$\phi_{l}$} & {Parameter of multinomial distribution for index $i_{t}$ of the position distribution} \\ 
{$\theta_{l}$} & {Parameter of multinomial distribution for the image feature} \\ 
{$W_{l}$} & {Parameter of multinomial distribution for the names of places} \\ 
$LM$ & Language model (word dictionary) \\ 
$AM$ & Acoustic model for speech recognition \\ 
$\alpha $, $\beta$, $\gamma$, $\chi$, $\lambda$,& {Hyperparameters of prior distributions}\\$m_{0}$, $\kappa_{0}$, $V_{0}$, $\nu_{0}$& \\ \hline
\end{tabular}
\end{center}
  \end{minipage}
\end{center}
  \caption{Left: Graphical model representation of SpCoSLAM \citep{ataniguchi_IROS2017}. Gray nodes indicate observation variables, and white nodes are unobserved latent variables. Right: Description of random variables in SpCoSLAM.} 
  \label{fig:gmodel}
\end{figure*}

\subsection{Improvement of online learning based on particle filters in unsupervised Bayesian models}
\label{sec:RelatedWork:online}
As an approach involving Bayesian models that is similar to our model, there are related studies on object concepts.
In particular, \citet{araki2012online_IROS} proposed online Multimodal Latent Dirichlet Allocation (oMLDA) to acquire object concepts in an online manner, and combined this with the Nested Pitman--Yor Language Model (NPYLM), making it possible to perform lexical acquisition of unknown words sequentially. 
\citet{aoki2016online} constructed an algorithm that can infer an approximately global optimal solution by representing it as a single integrated model. 
The NPYLM is an unsupervised morphological analysis method based on a statistical model that enables word segmentation exclusively from phoneme sequences \citep{mochihashi2009bayesian}.
In addition, \citet{nishihara2017online} was able to reduce phoneme recognition errors by applying PFoMDLA to inferences using a particle filter instead of oMLDA.
In these studies, online learning was realized as an algorithm in unsupervised machine learning.
A spatial concept requires more real-time processing than an object concept because the robot learns spatial concepts while it moves through the environment.
The mobile robot should not halt its spatial movement for calculations.
Therefore, a more efficient and scalable algorithm is required.

\citet{canini2009online} improved the accuracy of an online algorithm based on a particle filter with the rejuvenation technique.
This technique resamples some randomly selected samples of previous observation data from a conditional probabilistic distribution similar to Gibbs sampling.
For a completely random choice, the robot needs to memorize all of the previous data.
Rejuvenation can deal with the problem of degenerating particles in particle filters.
In this study, we introduce rejuvenation into our SpCoSLAM online learning algorithm.
In our algorithm, we perform resampling from some recent data.
Therefore, we consider that it will be possible to improve the estimation accuracy efficiently.

As another particle filter approach, \citet{borschinger2011particle} proposed an online algorithm based on a Bayesian model for word segmentation.
In addition, \citet{borschinger2012using} presented an incremental learning algorithm that introduces rejuvenation to a particle filter.
They improved the performance of word segmentation with higher accuracy.
The studies above were premised on segmentation of sequences without phoneme recognition errors.
In this study, by contrast, the online word segmentation task is particularly challenging because phoneme recognition errors are included in speech recognition results.

\section{SpCoSLAM: Online learning for spatial concepts and lexical acquisition with mapping}
\label{sec:SpCoSLAM}
\subsection{Overview}
\label{sec:SpCoSLAM:overview}
SpCoSLAM has the advantage that spatial concept formation, lexical acquisition, and SLAM, can be performed simultaneously by an integrated model.
Figure~\ref{fig:gmodel} shows the graphical model of SpCoSLAM and lists each variable of the graphical model. 
The details of the formulation of the generation process represented by the graphical model are described in \citet{ataniguchi_IROS2017}.
The method learns sequential spatial concepts for unknown environments without maps. 
It also learns the many-to-many correspondences between places and words via spatial concepts and can mutually complement the uncertainty of information using multimodal information.
Furthermore, the proposed method estimates an appropriate number of clusters of spatial concepts and position distributions depending on the data by using the so-called online Chinese Restaurant Process (CRP) \citep{aldous1985exchangeability}, one of the constitutive methods of the Dirichlet Process (DP). 
In addition, lexical acquisition including unknown words is possible by sequentially updating the language model.

The procedure of SpCoSLAM for each step is described as follows.
(a) The robot obtains Weighted Finite-State Transducer (WFST) speech recognition results from the user's speech signals using a language model. %
(b) The robot obtains the likelihood of self-localization by performing FastSLAM. 
(c) The robot segments the WFST speech recognition results using an unsupervised word segmentation approach called latticelm \citep{neubig2012bayesian}. 
(d) The robot obtains the latent variables of spatial concepts by sampling.
(e) The robot obtains the marginal likelihood of the observed data as the importance weight. 
(f) The robot updates the environmental map. 
(g) The robot estimates the set of model parameters of the spatial concepts from the observed data and the sampled variables. 
(h) The robot updates the language model of the maximum weight for the next step. 
(i) The particles are resampled according to their weights. Steps (b) -- (g) are performed for each particle.

\subsection{Formulation of the online learning algorithm}
\label{sec:SpCoSLAM:learning}
Our previously proposed online learning algorithm, SpCoSLAM, introduces sequential equation updates to estimate the parameters of the spatial concepts into the formulation of a Rao-Blackwellized Particle Filter (RBPF) \citep{doucet2000rao} in the FastSLAM~2.0 algorithm{, which is landmark-based SLAM \citep{montemerlo2003fastslam}, and the technique~\citep{gridbasedfastslam2007} applied to grid-based SLAM in a similar manner to that in FastSLAM 2.0}.
The particle filter is advantageous in that parallel processing can be easily applied because each particle can be calculated independently.

In the formulation of SpCoSLAM, the joint posterior distribution can be factorized to the probability distributions of a language model $LM$, a map $m$, the set of model parameters of spatial concepts $\Theta = \{ {\mathbf W}, \mbox{\boldmath $\mu $}, \mbox{\boldmath $\Sigma$}, {\mathbf \theta}, {\mathbf \phi}, \pi \}$, the joint distribution of the self-position trajectory $x_{0:t}$, and the set of latent variables $\mathbf{C}_{1:t} = \{i_{1:t},C_{1:t},S_{1:t} \}$.
We describe the joint posterior distribution of SpCoSLAM as follows:
\begin{eqnarray}
&&p(x_{0:t},\mathbf{C}_{1:t}, LM, \Theta, m 
\mid u_{1:t}, z_{1:t}, y_{1:t}, f_{1:t}, AM, \mathbf{h}) \nonumber \\
&&=p(LM \mid S_{1:t}, \lambda)
p(\Theta \mid x_{0:t}, \mathbf{C}_{1:t}, f_{1:t}, \mathbf{h})p(m \mid x_{0:t}, z_{1:t}) \nonumber \\
&&\hspace{1.0em}\cdot~\underbrace{p(x_{0:t},\mathbf{C}_{1:t} \mid u_{1:t}, z_{1:t}, y_{1:t}, f_{1:t}, AM, \mathbf{h})}_{\rm Particle~filter}
\label{eq:spcoslam}
\end{eqnarray}
where the set of hyperparameters is denoted by $\mathbf{h}= \{ \alpha,\beta,\gamma,\chi,\lambda, m_{0},\kappa_{0}, V_{0},\nu_{0} \}$.
It is noteworthy that the speech signal $y_{t}$ is not observed during all time-steps.
Herein, the proposed method is equivalent to FastSLAM 2.0 when $y_{t}$ is not observed, i.e., when the speech signal is a trigger for the place categorization.

\subsubsection{{Particle filter algorithm}}
The particle filter algorithm uses Sampling Importance Resampling (SIR).
The importance weight is denoted by $\omega_{t}^{[r]}={P_{t}^{[r]}}/{Q_{t}^{[r]}}$ for each particle, 
where $r$ is the particle index. 
{The target distribution is $P_{t}^{[r]}$, and the proposal distribution is $Q_{t}^{[r]}$.}
The number of particles is $R$.
The following equations are also calculated for each particle $r$; however, the subscripts representing the particle index are omitted.

{We apply two modifications related to the weighting of the original SpCoSLAM algorithm \citep{ataniguchi_IROS2017}: (i) additional weight for $i_{t}$, $C_{t}$, and $x_{t}$ (AW), and (ii) weight for selecting a language model $LM$ (WS).
These modifications are more theoretically reasonable than the original SpCoSLAM model, and our proposed SpCoSLAM 2.0 online learning algorithm is extended on their basis.}

{We describe the target distribution $P_{t}$ that modified the derivation of \citet{ataniguchi_IROS2017} as follows:}
\begin{eqnarray}
P_{t} &=&p(x_{0:t},\mathbf{C}_{1:t} \mid u_{1:t}, z_{1:t}, y_{1:t}, f_{1:t}, AM, \mathbf{h}) \nonumber \\
&\approx & p(i_{t},C_{t} \mid x_{0:t}, i_{1:t-1},C_{1:t-1},S_{1:t},f_{1:t},\mathbf{h}) \nonumber \\
&&\hspace{0.0em}\cdot~p(z_{t} \mid x_{t}, m_{t-1})p(f_{t} \mid C_{1:t-1}, f_{1:t-1}, \mathbf{h}) \nonumber \\
&&\hspace{0.0em}\cdot~p(x_{t} \mid x_{t-1}, u_{t})p(S_{t} \mid S_{1:t-1},y_{1:t},AM,\lambda)\nonumber \\
&&\hspace{0.0em}\cdot~\underbrace{p( x_{t} \mid x_{0:t-1}, i_{1:t-1},{C}_{1:t-1}, \mathbf{h})}_{{\rm Additional~part}} \nonumber \\
&&\hspace{0.0em}\cdot~\frac{p(S_{t} \mid S_{1:t-1}, C_{1:t-1}, \alpha,\beta)}{p(S_{t} \mid S_{1:t-1}, \beta)} \cdot P_{t-1}, 
\label{eq:spcoslam_target}
\end{eqnarray}
{where the term $p( x_{t} \mid x_{0:t-1}, i_{1:t-1},{C}_{1:t-1}, \mathbf{h})$ is the additional part compared to the original equation.}

{Here,  
the target distribution for the particle filter is the marginal joint posterior distribution of the self-positions $x_{0:t}$ and the set of latent variables $\mathbf{C}_{1:t}$ because it is based on the RBPF technique adopted in FastSLAM in the same manner.
The latent variables that are local parameters are estimated by a particle filter, and the probability distributions for global parameters $LM$, $\Theta$, and $m$ are calculated and held independently for each estimated particle.}

We describe the proposal distribution $Q_{t}$ as follows:
\begin{eqnarray}
Q_{t}&=&q(x_{0:t},\mathbf{C}_{1:t} \mid u_{1:t}, z_{1:t}, y_{1:t}, f_{1:t}, AM, \mathbf{h}) \nonumber \\
&=&p(x_{t} \mid x_{t-1},z_{t},m_{t-1},u_{t}) \nonumber \\
&&\cdot~p(i_{t},C_{t} \mid x_{0:t}, i_{1:t-1},C_{1:t-1},S_{1:t},f_{1:t},\mathbf{h}) \nonumber \\
&&\cdot~p(S_{t} \mid S_{1:t-1},y_{1:t},AM,\lambda) \cdot Q_{t-1}. 
\label{eq:spcoslam_proposal}
\end{eqnarray}

Then, $p(x_{t} \mid x_{t-1},z_{t},m_{t-1},u_{t})$ is equivalent to the proposal distribution of FastSLAM~2.0.
The probability distribution of $i_{t}$ and $C_{t}$ is the marginal distribution pertaining to the set of model parameters $\Theta$.
This distribution can be calculated using a formula equivalent to collapsed Gibbs sampling.
The details are described in \citet{ataniguchi_IROS2017}.

\subsubsection{{Sampling of words using speech recognition and word segmentation}}
We approximate the probability distribution of $S_{t}$ in (\ref{eq:spcoslam_proposal}) as speech recognition with the language model $LM_{t-1}$ and unsupervised word segmentation using the WFST speech recognition results with latticelm \citep{neubig2012bayesian} as follows:
\begin{eqnarray}
&&p(S_{t} \mid S_{1:t-1},y_{1:t},AM,\lambda) \nonumber \\
&&\approx {\rm latticelm}(S_{1:t} \mid {\cal L}_{1:t},\lambda){\rm SR}({\cal L}_{1:t} \mid y_{1:t},AM,LM_{t-1}) \nonumber \\
\label{eq:spcoslam_latticelm_SR}
\end{eqnarray}
where ${\rm SR}()$ denotes the function of speech recognition, ${\cal L}_{1:t}$ denotes the speech recognition results in WFST format, which is a word graph representing the speech recognition results.
In the original mathematical formulas, only $S_{t}$ should be obtained by sampling.
However, latticelm is a tool originally designed for batch learning. 
In addition, in order to perform unsupervised word segmentation, it is necessary to extract statistical information from the observation data. 
Therefore, resampling is necessary using all data from 1 to $t$, instead of exclusively using the distribution at time-step~$t$.

\subsubsection{Additional weight for $i_{t}$, $C_{t}$, and $x_{t}$ (AW)}
\label{sec:SpCoSLAM:modification:AW}
{Finally, the importance weight $\omega_{t}$ modified from \citet{ataniguchi_IROS2017} is represented as follows:}
\begin{eqnarray}
\omega_{t} 
&\approx& 
\sum_{i_{t}=k} \Bigl[ p(x_{t} \mid x_{0:t-1}, i_{1:t-1}, i_{t}=k, \mathbf{h}) \nonumber \\
&&~\underbrace{\qquad \cdot~\sum_{C_{t}=l} p( i_{t}=k, C_{t}=l \mid C_{1:t-1}, i_{1:t-1}, \mathbf{h}) \Bigr]}_{{\rm Additional~part}} \nonumber \\
&&\cdot~p(z_{t} \mid m_{t-1}, x_{t-1},u_{t})p(f_{t} \mid C_{1:t-1}, f_{1:t-1}, \mathbf{h}) \nonumber \\
&&\cdot~\frac{p(S_{t} \mid S_{1:t-1}, C_{1:t-1}, \alpha,\beta)}{p(S_{t} \mid S_{1:t-1}, \beta)} %
\cdot \omega_{t-1}.
\label{eq:spcoslam_weight}
\end{eqnarray}

\newsavebox{\boxa} \sbox{\boxa}{\ref{eq:spcoslam_weight}}

Unlike the original SpCoSLAM algorithm, the marginal likelihood for $i_{t}$ and $C_{t}$ weighted by the marginal likelihood for the position distribution was added to {the additional part of the first term on the right side of ({\usebox\boxa})}.
The amount of calculations does not increase because most of the formulas for weight $\omega_{t}$ are already calculated when $i_{t}$ and $C_{t}$ are sampled. 
Weight calculation in consideration of the likelihood of the entire model can be realized by (\ref{eq:spcoslam_weight}).
This is described in Algorithm~\ref{alg:SpCoSLAM2_2} (Line 16) and Algorithm~\ref{alg:SpCoSLAM2_scalable} (Line~17).

\subsubsection{Weight for selecting a language model $LM$ (WS)}
\label{sec:SpCoSLAM:modification:WS}

In the formulation of (\ref{eq:spcoslam}), it is desirable to estimate the language model $LM_{t}$ for each particle.
In other words, speech recognition of the amount of data multiplied by the number of particles for each teaching utterance must be performed.
In this paper, to reduce the computational cost, we use a language model $LM_{t}$ of a particle with the maximum weight for speech recognition.

We also modify the weight for selecting the language model from the entire weight $\omega_{t}$ of the model to the weight $\omega_{S}$ related to word information:
\begin{eqnarray}
\omega_{S} = \frac{p(S_{1:t} \mid C_{1:t-1}, \alpha,\beta)}{p(S_{1:t} \mid \beta)}.
\label{eq:spcoslam2_0_WS}
\end{eqnarray}
The segmentation result from all of the uttered sentences for each particle changes at every step because the word segmentation processes use all previous data.
Indeed, better word segmentation results can be selected by a weight that considers not only current data but also previous data.
In addition, this modified weight corresponds to mutual information used for selecting the word segmentation results in SpCoA++ \citep{taniguchi2018unsupervised}.
This is described in Algorithm~\ref{alg:SpCoSLAM2_2} (Line 23) and Algorithm~\ref{alg:SpCoSLAM2_scalable} (Line 24).

\section{SpCoSLAM 2.0: Improved and scalable online learning algorithm}
\label{sec:proposed} 

In this section, we describe an improved and scalable online learning algorithm, SpCoSLAM 2.0, that overcomes the problems in the original algorithm.
Although the generative process and graphical model for SpCoSLAM are the same, the learning algorithm is different.
SpCoSLAM 2.0 is a novel learning algorithm proposed with a modified mathematical formulation that retains the model structure, similar to the extension from FastSLAM to FastSLAM 2.0.
First, the algorithm is improved by introducing techniques such as rejuvenation, as explained in Section~\ref{sec:proposed:improvement}. 
Next, a scalable algorithm is developed to reduce the calculation time while maintaining higher accuracy than the original algorithm, as described in Section~\ref{sec:proposed:scalability}.

\begin{figure*}[!tb]
  \begin{center}
    \includegraphics[width=0.88\hsize]{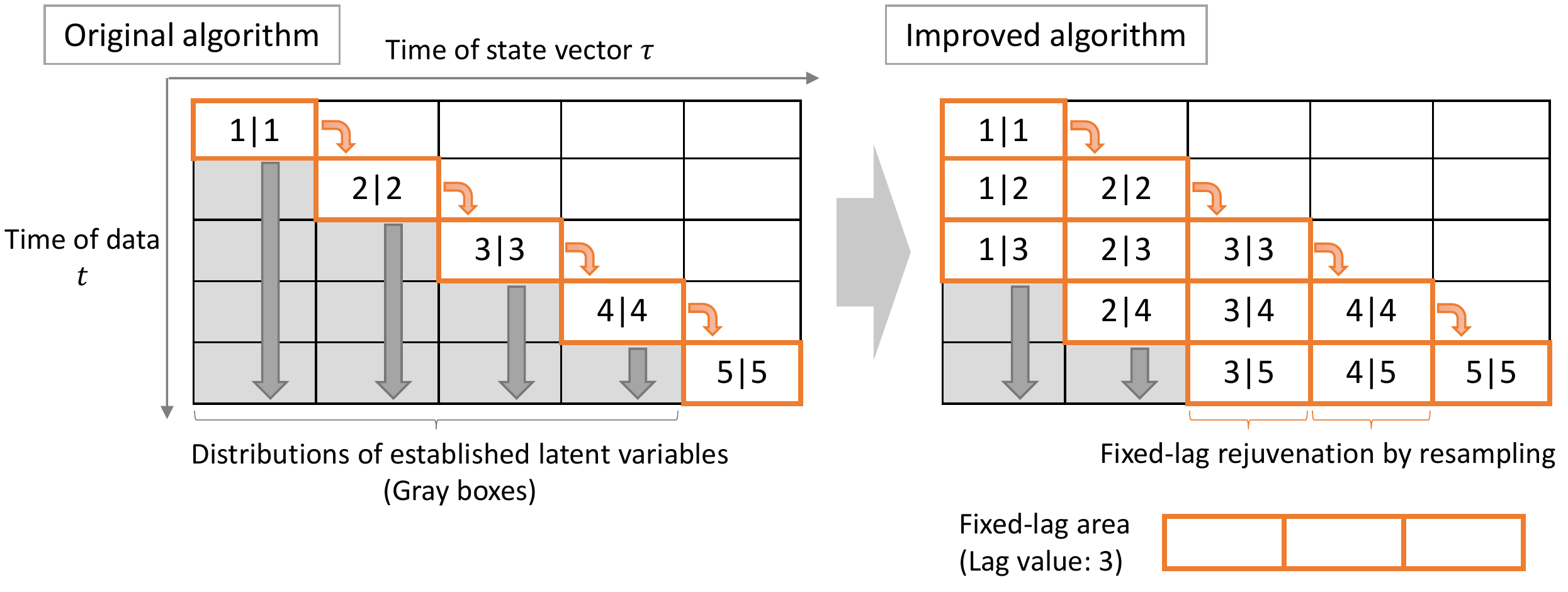}
    \caption{Overview of the Fixed-Lag Rejuvenation of $i_{t}$ and $C_{t}$. Left: Naive online learning in the original algorithm. Right: Online learning using FLR in the improved algorithm. The thick orange frame is estimated by sampling. In this case, the fixed-lag value $T_{L}$ is three. The gray boxes mean that the estimated value will never again be updated, i.e., distributions of already immobilized (fixed) latent variables by online learning.}
    \label{fig:FLR_improved}
  \end{center}
\end{figure*}

\begin{algorithm}[!tb]  
\caption{SpCoSLAM 2.0: Improved algorithm} 
\label{alg:SpCoSLAM2_2} 
\begin{algorithmic}[1]
\Procedure{${\rm SpCoSLAM2.0}$}{$X_{t-1},u_{t},z_{t},f_{1:t},y_{1:t}$}
\State $\bar{X}_{t} = X_{t} = \emptyset$
\State ${\cal L}_{1:t} = {\rm SR}({\cal L}_{1:t} \mid y_{1:t},AM,LM_{t-1})$
\For{$r=1$ to $R$}
\State $\acute{x}_{t}^{[r]} = {\rm\bf sample\_motion\_model}(u_{t},x_{t-1}^{[r]})$
\State ${x}_{t}^{[r]} = {\rm\bf scan\_matching}(z_{t}, \acute{x}_{t}^{[r]}, m_{t-1}^{[r]})$ 
\For{$j=1$ to $J$}
\State ${x}_{j} = {\rm\bf sample\_motion\_model}(u_{t},x_{t-1}^{[r]})$
\EndFor
\State $\omega_{z}^{[r]} = \displaystyle \sum_{j=1}^{J} {\rm\bf measurement\_model}(z_{t}, {x}_{j}, m_{t-1}^{[r]})$ 
\State $S_{1:t}^{[r]} \sim {\rm latticelm}(S_{1:t} \mid {\cal L}_{1:t},\lambda)$
\For{{$\tau = t-T_{L}+1$ to $t$}}
\State {$i_{\tau}^{[r]},C_{\tau}^{[r]} \sim p(i_{\tau},C_{\tau} \mid x_{0:t}^{[r]},S_{1:t}^{[r]},f_{1:t},$ }
\Statex \hspace{14.0em}{$ i_{ \{1:t\mid \lnot \tau \} }^{[r]},C_{ \{1:t\mid \lnot \tau \} }^{[r]},\mathbf{h})$ }
\EndFor
\State $\omega_{f}^{[r]} = p(f_{t} \mid C_{1:t-1}^{[r]}, f_{1:t-1}, \alpha,\chi)$
\State {$\omega_{ic}^{[r]}\,=\,\displaystyle \sum_{i_{t}=k}\,\Bigl[ p(x_{t}^{[r]} \mid x_{0:t-1}^{[r]}, i_{1:t-1}^{[r]}, i_{t}=k, \mathbf{h})$}
\Statex {\hspace{3.0em}$\cdot \displaystyle \sum_{C_{t}=l} p( i_{t}=k, C_{t}=l\,\mid\,C_{1:t-1}^{[r]}, i_{1:t-1}^{[r]}, \mathbf{h}) \Bigr]$}
\State $\omega_{s}^{[r]} = \displaystyle \frac{p(S_{t}^{[r]}\,|\,S_{1:t-1}^{[r]}, C_{1:t-1}^{[r]}, \alpha,\beta)}{p(S_{t}^{[r]} \mid S_{1:t-1}^{[r]}, \beta)} $
\State $\omega_{t}^{[r]} = \omega_{z}^{[r]} \cdot \omega_{f}^{[r]} \cdot \omega_{s}^{[r]} \cdot \omega_{ic}^{[r]}$
\State $m_{t}^{[r]} = {\rm\bf updated\_occupancy\_grid}(z_{t},x_{t}^{[r]},m_{t-1}^{[r]})$
\State $\Theta_{t}^{[r]} = {E}[p(\Theta \mid x_{0:t}^{[r]}, \mathbf{C}_{1:t}^{[r]}, f_{1:t}, \mathbf{h})]$
\State $\bar{X}_{t} = \bar{X}_{t} \cup \langle x_{0:t}^{[r]}, \mathbf{C}_{1:t}^{[r]}, m_{t}^{[r]}, \Theta_{t}^{[r]}, \omega_{t}^{[r]} \rangle$
\EndFor
\State $S_{1:t}^{*} = \displaystyle \argmax_{S_{1:t}^{[r]}} \sum_{r=1}^{R} \omega_{S}^{[r]} \delta(S_{1:t} - S_{1:t}^{[r]})$
\State $LM_{t} \sim {\rm NPYLM}(LM \mid S_{1:t}^{*}, \lambda)$ 
\For{$r=1$ to $R$}
\State draw $i$ with probability $\propto \omega_{t}^{[i]}$ 
\State add $\langle x_{0:t}^{[i]}, \mathbf{C}_{1:t}^{[i]}, m_{t}^{[i]}, \Theta_{t}^{[i]}, LM_{t} \rangle$ to $X_{t}$
\EndFor
\Return $X_{t}$
\EndProcedure
\end{algorithmic}
\end{algorithm}

\subsection{Improving the estimation accuracy}
\label{sec:proposed:improvement} 
We now turn to the details of the improved algorithm.
Here, we introduce two elements: fixed-lag rejuvenation of latent variables, and re-segmentation of word sequences.
A pseudo-code for the improved algorithm is given in Algorithm~\ref{alg:SpCoSLAM2_2}.

\subsubsection{Fixed-lag rejuvenation of $i_{t}$ and $C_{t}$ (FLR--$i_{t}$, $C_{t}$)}
\label{sec:proposed:improvement:FLR} 

\citet{canini2009online} demonstrated improved accuracy with rejuvenation by resampling previous samples randomly.
{This is based on a result of the independent and identically distributed (i.i.d.) assumption on the latent variables in the Latent Dirichlet Allocation (LDA) model.}
However, {in the case of selecting from previous data of all time points,}
all previous samples need to be held in the memory.
In the proposed algorithm, we introduce Fixed-Lag Rejuvenation (FLR) inspired by the Monte Carlo fixed-lag smoother \citep{kitagawa2014computational}.
{This approach is similar to the sampling strategy of fixed-lag roughening for particle filter-based SLAM in \citet{beevers2007fixed}. 
\citet{beevers2007fixed} indicated that the statistical estimation error could be reduced by applying Markov Chain Monte Carlo (MCMC)--based sampling to the trajectory samples over a fixed lag at each time step.}

The fixed-lag smoother is a particle smoothing method that estimates particles approximating the smoothing distribution $p(\mathbf{C}_{\tau} \mid D_{1:t})$ $(\tau<t)$, where $D$ is observed data.
It is obtained by a simple modification to the particle filter.
In this algorithm, particles are saved from time-step $t - T_{L} + 1$ to $t$ and are resampled according to the weight based on newly observed data each step.
Here, the value of the fixed-lag is denoted by $T_{L}$.
This technique means that the particles at step $\tau$ can be estimated not by using the observed data $D_{1:\tau}$, but rather with $D_{1:\tau + T_{L}}$, i.e., the smoothing distribution $p(\mathbf{C}_{\tau} \mid D_{1:\tau + T_{L}})$.
In general, a smoothing method {such as a fixed-lag particle smoother} provides more accurate estimations than naive online estimation methods such as a particle filter {in estimating the joint posterior distribution of latent variables}. 

Figure~\ref{fig:FLR_improved} shows an overview of the FLR of $i_{t}$ and $C_{t}$.
The notation $\tau \mid t$ in the box in Fig.~\ref{fig:FLR_improved} is shorthand notation for the subscript representing the time-step in the conditional marginal posterior distribution, e.g., $p(\mathbf{C}_{\tau} \mid D_{1:t})$. 
The FLR is the process of sampling the latent variables $i_{\tau}$ and $C_{\tau}$ by iterating $T_{L}$ times from the previous step $t-T_{L}+1$ to the current step $t$ for each particle as follows:
\begin{eqnarray}
i_{\tau},C_{\tau} \sim p(i_{\tau},C_{\tau} \mid x_{0:t},S_{1:t},f_{1:t}, i_{ \{1:t\mid \lnot \tau \} },C_{ \{1:t\mid \lnot \tau \} },\mathbf{h})
\nonumber \\
\label{eq:spcoslam2_0_FLR}
\end{eqnarray}
where $i_{ \{1:t\mid \lnot \tau \} }$ and $C_{ \{1:t\mid \lnot \tau \} }$ denote sets of elements from $1$ to $t$ without the elements of step $\tau$.
In this case, the latent variables of step $t-T_{L}$ can be sampled using data up to step $t$, as described in Algorithm~\ref{alg:SpCoSLAM2_2} (Lines 12--14).
Equation (\ref{eq:spcoslam2_0_FLR}) is the same as the conditional posterior probability distribution for marginalized (collapsed) Gibbs sampling used in batch learning.
Therefore, the FLR corresponds to slightly iterate Gibbs sampling for some recent previous latent variables in online learning.

\subsubsection{Re-segmentation of word sequences~(RS)}
\label{sec:proposed:improvement:RS} 
We introduce re-segmentation of word sequences to improve the accuracy of word segmentation.
In the original algorithm, we approximated the left side of (\ref{eq:spcoslam_latticelm_SR}) by registering the word sequences segmented by latticelm to the word dictionary.
However, this can be considered a process of sampling a language model $LM$ from word sequences $S_{1:t}^{*}$ and a hyperparameter $\lambda$ of a language model.
Therefore, we adopt NPYLM, an unsupervised word segmentation method  \citep{mochihashi2009bayesian}, to estimate a language model from the word sequences as follows:
\begin{eqnarray}
LM &\sim& 
{\rm NPYLM}(LM \mid S_{1:t}^{*}, \lambda). 
\label{eq:spcoslam_latticelm_LM}
\end{eqnarray}

The procedure of introducing the RS is as follows:
(i) word sequences $S_{1:t}$ are obtained by WFST speech recognition and latticelm; (ii) word sentences $S_{1:t}^{*}$ of a maximum likelihood particle are converted into syllable sequences, and segmented into word sequences using NPYLM; (iii) the word dictionary $LM$ is updated using segmented words, as described in Algorithm~\ref{alg:SpCoSLAM2_2} (Line 24).
In this manner, we can overcome problematic words that tend to become under-segmented while taking into account the uncertainty of speech recognition errors by latticelm.
Note that there is a discrepancy between the words used for spatial concept acquisition and the word set registered in the word dictionary.

\begin{algorithm}[tb] 
\caption{SpCoSLAM 2.0: Scalable algorithm} 
\label{alg:SpCoSLAM2_scalable} 
\begin{algorithmic}[1]
\Procedure{${\rm SpCoSLAM2.0}$}{$X_{t-1},u_{t},z_{t},{f_{t'+1:t},y_{t'+1:t}}$}
\State $\bar{X}_{t} = X_{t} = \emptyset$
\State {$t' = t-T_{L}$}
\State {${\cal L}_{t'+1:t} = {\rm SR}({\cal L}_{t'+1:t} \mid y_{t'+1:t},AM,LM_{t'})$}
\For{$r=1$ to $R$}
\State $\acute{x}_{t}^{[r]} = {\rm\bf sample\_motion\_model}(u_{t},x_{t-1}^{[r]})$
\State ${x}_{t}^{[r]} = {\rm\bf scan\_matching}(z_{t}, \acute{x}_{t}^{[r]}, m_{t-1}^{[r]})$ 
\For{$j=1$ to $J$}
\State ${x}_{j} = {\rm\bf sample\_motion\_model}(u_{t},x_{t-1}^{[r]})$
\EndFor
\State $\omega_{z}^{[r]} = \displaystyle \sum_{j=1}^{J} {\rm\bf measurement\_model}(z_{t}, {x}_{j}, m_{t-1}^{[r]})$ 
\State {$S_{t'+1:t}^{[r]} \sim {\rm latticelm}(S_{t'+1:t} \mid {\cal L}_{t'+1:t},\lambda)$}
\For{{$\tau = t'+1$ to $t$} }
\State {$i_{\tau}^{[r]},C_{\tau}^{[r]} \sim p(i_{\tau},C_{\tau} \mid x_{t'+1:t}^{[r]},S_{t+1:t}^{[r]},f_{t'+1:t},$ }
\Statex \hspace{10.0em}{$ i_{ \{t'+1:t\mid \lnot \tau \} }^{[r]},C_{ \{t'+1:t\mid \lnot \tau \} }^{[r]},H_{t'}^{[r]})$}
\EndFor
\State {$\omega_{f}^{[r]} = p(f_{t} \mid C_{t'+1:t-1}^{[r]}, f_{t'+1:t-1}, H_{t'}^{[r]})$}
\State {$\displaystyle \omega_{ic}^{[r]}=\sum_{i_{t}=k}\,\Bigl[ p(x_{t}^{[r]}\mid x_{t'+1:t-1}^{[r]}, i_{t'+1:t-1}^{[r]}, $}
\Statex {\hspace{9.0em}$i_{t}=k, H_{t'}^{[r]})\displaystyle \sum_{C_{t}=l} p( i_{t}=k, C_{t}=l\mid$} 
\Statex {\hspace{9.0em}$C_{t'+1:t-1}^{[r]}, i_{t'+1:t-1}^{[r]}, H_{t'}^{[r]}) \Bigr]$}
\State {$\omega_{s}^{[r]} = \displaystyle \frac{p(S_{t}^{[r]} \mid S_{t'+1:t-1}^{[r]}, C_{t'+1:t-1}^{[r]}, H_{t'}^{[r]})}{p(S_{t}^{[r]} \mid S_{t'+1:t-1}^{[r]}, H_{t'}^{[r]})}$}
\State $\omega_{t}^{[r]} = \omega_{z}^{[r]} \cdot \omega_{f}^{[r]} \cdot \omega_{s}^{[r]} \cdot \omega_{ic}^{[r]}$
\State $m_{t}^{[r]} = {\rm\bf updated\_occupancy\_grid}(z_{t},x_{t}^{[r]},m_{t-1}^{[r]})$
\State {$H_{t}^{[r]} = {F}[p(\Theta \mid x_{t'+1:t}^{[r]}, \mathbf{C}_{t'+1:t}^{[r]}, f_{t'+1:t}, H_{t'}^{[r]})]$ }
\State {$\bar{X}_{t} = \bar{X}_{t} \cup \langle x_{t'+1:t}^{[r]}, \mathbf{C}_{t'+1:t}^{[r]}, m_{t}^{[r]}, H_{t'+1:t}^{[r]}, \omega_{t}^{[r]} \rangle$}
\EndFor
\State {$S_{t'+1:t}^{*} = \displaystyle \argmax_{S_{t'+1:t}^{[r]}} \sum_{r=1}^{R} \omega_{S}^{[r]} \delta(S_{t'+1:t} - S_{t'+1:t}^{[r]})$}
\State {$LM_{t} = \displaystyle \argmax_{LM} p(LM \mid S_{t'+1:t}^{*}, LM_{t'}, \lambda)$} 
\For{$r=1$ to $R$}
\State draw $i$ with probability $\propto \omega_{t}^{[i]}$ 
\State {add $\langle x_{t'+1:t}^{[i]}, \mathbf{C}_{t'+1:t}^{[i]}, m_{t}^{[i]}, H_{t'+1:t}^{[i]}, LM_{t'+1:t} \rangle$ to $X_{t}$}
\EndFor
\Return $X_{t}$
\EndProcedure
\end{algorithmic}
\end{algorithm}

\begin{figure*}[!tb]
  \begin{center}
    \includegraphics[width=0.88\hsize]{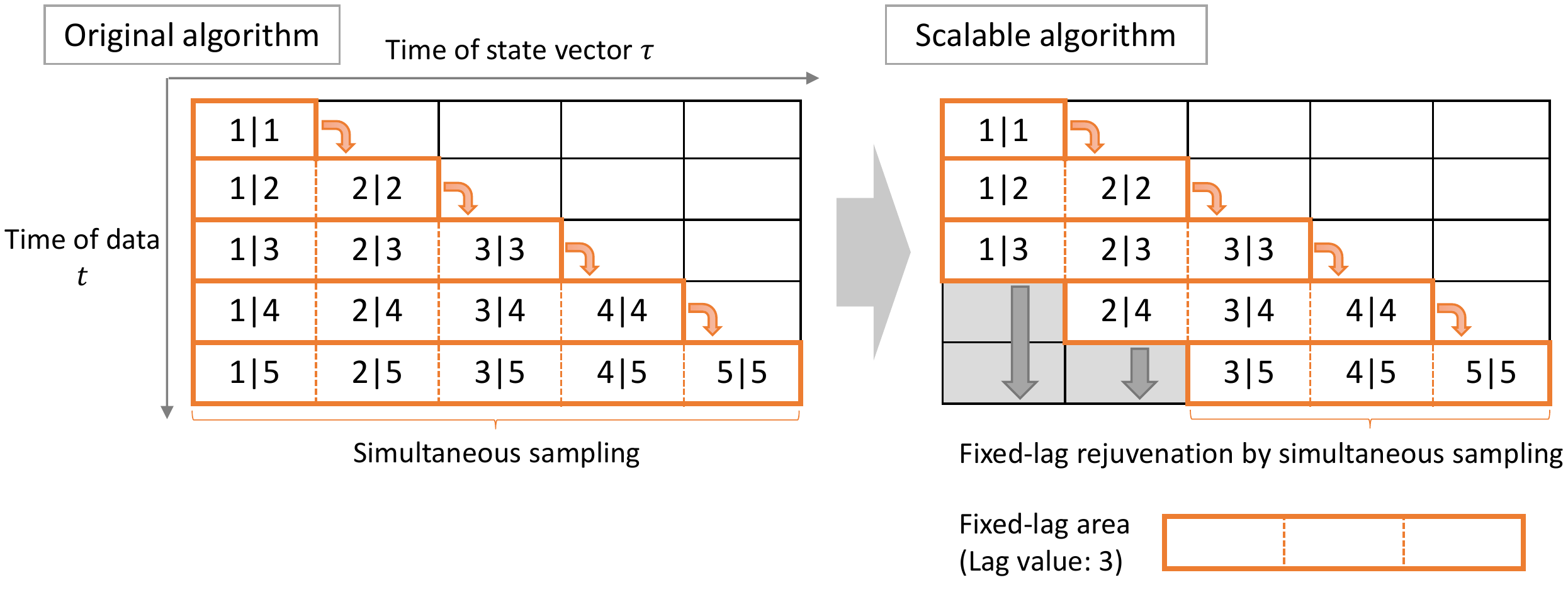}
    \caption{Overview of the Fixed-Lag Rejuvenation of $S_{t}$. Left: Batch learning with the original algorithm. Right: Pseudo-online learning using FLR in the scalable algorithm. The thick orange frame is estimated by sampling from the joint distribution. In this case, the fixed-lag value $T_{L}$ is three. The gray boxes denote that the estimated value will never be updated again, i.e., distributions of already immobilized (fixed) latent variables by online learning.}
    \label{fig:FLR_scalable}
  \end{center}
\end{figure*}

\subsection{Scalability for reduced computational cost}
\label{sec:proposed:scalability} 
In this section, we describe the details of the scalable algorithm.
Here, we introduce two elements: the sequential Bayesian update of the parameters in the posterior distribution, and unsupervised word segmentation from WFST speech recognition results using FLR.
The scalable algorithm can be combined with the FLR $C_{t}$, $i_{t}$ of the improved algorithm.
The pseudo-code for the scalable algorithm is given in Algorithm~\ref{alg:SpCoSLAM2_scalable}.

\subsubsection{Sequential Bayesian update of parameters in the posterior distribution (SBU)}
\label{sec:proposed:scalability:SBU} 

We introduce a Sequential Bayesian Update (SBU) for the posterior hyperparameters {$H_{t}$} in the posterior distribution.
In the original algorithm, the model parameters $\Theta$ are estimated from all the data $D_{1:t}=\{ f_{1:t}, y_{1:t} \}$ and the set of latent variables $\mathbf{C}_{1:t}$ during each step.
However, FastSLAM avoids holding all the previous data by updating a map $m_{t}$ from $x_{t}$, $z_{t}$, and $m_{t-1}$ sequentially. That is, it assumes the measurement model $p(z_{t} \mid x_{0:t}, z_{1:t-1})=p(z_{t} \mid x_{t}, m_{t-1})$ and the updated occupancy grid map $p(m_{t} \mid x_{0:t},z_{1:t})=p(m_{t} \mid x_{t},z_{t},m_{t-1})$.
Similarly, the posterior hyperparameters $H_{t}$ can be calculated from the new data ${D}_{t}$, latent variables $\mathbf{C}_{t}$, and posterior hyperparameters $H_{t-1}$ from previous steps.
Thus, both the computational and memory efficiency, crucial for long-term learning with real robots, can be significantly improved.
The SBU for the posterior hyperparameters is calculated as follows:
\begin{eqnarray}
p(\Theta \mid H_{t}) 
&=&p(\Theta \mid D_{1:t},\mathbf{C}_{1:t}, \mathbf{h}) \nonumber \\
&=&p(\Theta \mid D_{t},\mathbf{C}_{t}, \{ D_{1:t-1},\mathbf{C}_{1:t-1}, \mathbf{h} \} ) \nonumber \\
&=&p(\Theta \mid D_{t},\mathbf{C}_{t}, H_{t-1}) \nonumber \\
&\propto &p(D_{t} \mid \mathbf{C}_{t}, \Theta)p(\Theta \mid H_{t-1}). 
\label{eq:spcoslam_hyperupdate}
\end{eqnarray}
These posterior hyperparameters $H_{t}$ can also be used to sample $\mathbf{C}_{t}$.
In the implementation, it suffices to hold values of the statistics obtained during the calculation of the posterior distribution. 
Here, the calculation results from the left side and the right side of (\ref{eq:spcoslam_hyperupdate}) are strictly the same.
{The SBU approach is also said to keep track of sufficient statistics in the particle filter \citep{kantas2015particle}.}

The SBU equation is used together with FLR as follows:
\begin{eqnarray}
p(\Theta \mid H_{t}) 
&=&p(\Theta \mid D_{1:t},\mathbf{C}_{1:t}, \mathbf{h}) \nonumber \\
&=&p(\Theta \mid D_{t'+1:t},\mathbf{C}_{t'+1:t}, \{ D_{1:t'}, \mathbf{C}_{1:t'}, \mathbf{h} \} )  \nonumber \\
&=&p(\Theta \mid D_{t'+1:t},\mathbf{C}_{t'+1:t}, H_{t'}) \nonumber \\
&\propto &p(D_{t'+1:t} \mid \mathbf{C}_{t'+1:t}, \Theta)p(\Theta \mid H_{t'}), 
\label{eq:spcoslam_hyperupdate_lag}
\end{eqnarray}
where a time-step before the lag value is $t' = t - T_{L}$.
In this case, it is only necessary to hold the observed data and posterior hyperparameters of the number corresponding to the lag value $T_{L}$.
Equation~(\ref{eq:spcoslam_hyperupdate_lag}) is applied to Algorithm~\ref{alg:SpCoSLAM2_scalable}.

\begin{table}[tb]
\begin{center}
\caption{Computational complexity of the learning algorithms}
\begin{tabular}{p{140pt}l} \hline
Algorithm & Order \\ \hline
SpCpSLAM \citep{ataniguchi_IROS2017} & $O(NR)$ \\ 
SpCoSLAM 2.0 (Improved) & $O(NR)$\\ 
SpCoSLAM 2.0 (Scalable) & $O(T_{L}R)$ \\ \hline
SpCoA \citep{taniguchi_spcoa} (Batch learning) & $O(NG)$ \\ 
SpCoA++ \citep{taniguchi2018unsupervised} (Batch learning) & $O(NGMI)$ \\ \hline
\end{tabular}
\label{table:order}
\end{center}
\end{table}

\subsubsection{WFST speech recognition and unsupervised word segmentation using FLR (FLR--$S_{t}$)}
\label{sec:proposed:scalability:FLR} 
We describe the proposed algorithm that combines FLR and SBU to address problems of the unsupervised online word segmentation and to reduce the computation time simultaneously. 
FLR can also be extended to the sampling of $S_{t}$ in a pseudo-online manner.
Figure~\ref{fig:FLR_scalable} shows an overview of the FLR of $S_{t}$.
The notation $\tau \mid t$ takes the same meaning as it does in Fig.~\ref{fig:FLR_improved}.
The data used for speech recognition and word segmentation is modified from that in (\ref{eq:spcoslam_latticelm_SR}) to data with a fixed-lag interval.
In addition, speech recognition is performed using the initial syllable dictionary in the steps before step $T_{L}$ and using a word dictionary from step $t'$ in the steps proceeding step $T_{L}+1$.
In this case, we can perform word segmentation based on the statistical information collected from the WFSTs recognized using the number of data for the lag value $T_ {L}$.
FLR performs simultaneous sampling of word sequences $S_{t'+1:t}$ of time-steps from $t'+1$ to the current step $t$ as follows:
\begin{eqnarray}
S_{t'+1:t}&\sim &
p(S_{t'+1:t} \mid y_{t'+1:t},AM,S_{1:t'},\lambda) \nonumber \\
&&\approx {\rm latticelm}(S_{t'+1:t} \mid {\cal L}_{t'+1:t},\lambda) \nonumber \\
&&\quad \cdot~{\rm SR}({\cal L}_{t'+1:t} \mid y_{t'+1:t},AM,LM_{t'}). 
\label{eq:spcoslam_latticelm_FLRs}
\end{eqnarray}
Therefore, this approach can address the problem in the original algorithm by which incorrect word segmentation in early learning stages was propagated to the following learning stages.

Here, the amount of calculations is constant throughout each step, irrespective of the total amount of data.
This property of the FLR of $S_{t}$ is an important advantage in scalability.
However, 
there is a concern that word segmentation using FLR becomes inaccurate compared to batch learning because of the limited availability of statistical information.
Essentially, the scalable algorithm is a trade-off between calculation time and word segmentation accuracy.
In the language model update, the word dictionary $LM_{t}$ holds information regarding words $S_{t'+1:t}$ segmented from steps $t'+1$ to $t$ and the previous word dictionary $LM_{t'}$.
This is described in Algorithm~\ref{alg:SpCoSLAM2_scalable} (Lines 4, 12, and 25).

Table~\ref{table:order} shows the order of computational complexity for each learning algorithm.
The data number is denoted $N$, the number of particles $R$, the value of fixed-lag $T_{L}$, the number of iterations for Gibbs sampling in batch learning $G$, the number of candidates of word segmentation results for updating the language model in SpCoA++ $M$, and the number of iterations for the parameter estimation in SpCoA++ $I$.
Variables without $N$ are constants that can be preset by the user.
Among these algorithms, therefore, only the scalable algorithm does not depend on the number of data $N$.
In this case, the computational efficiency of the scalable algorithm is better than the original SpCoSLAM algorithm when $T_{L}<N$.

\begin{figure*}[tb]
  \begin{center}
    \includegraphics[width=1.00\hsize]{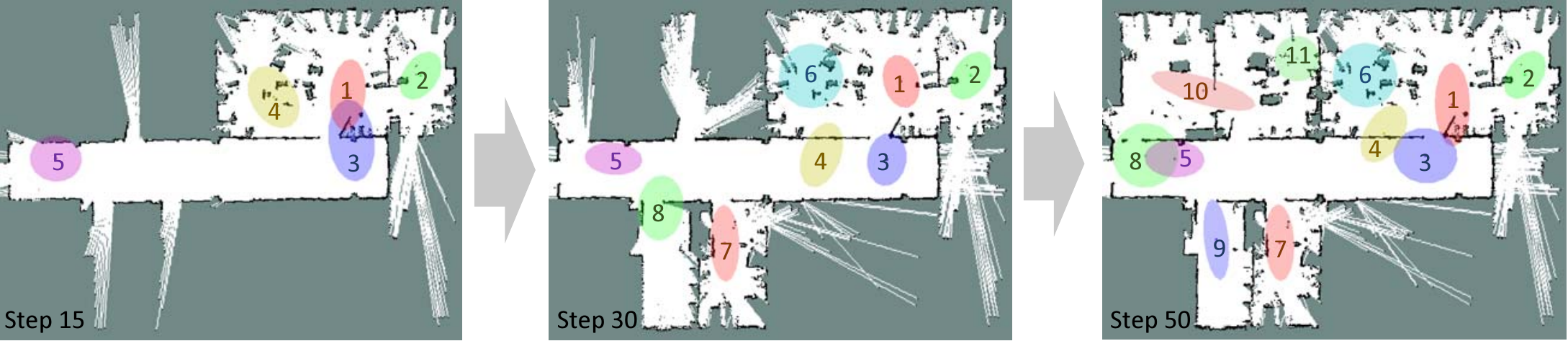}
    \\ 
    \vspace{10pt} 
\begin{tabular}{llllllll} \hline
Image&Correct word& 
$i_{t}$& Step 15 & 
$i_{t}$& Step 30 & 
$i_{t}$& Step 50
\\ 
&({\it English}) & 
& & & & & 
\\ \hline 
    \begin{minipage}{58pt}
    \centering
    \vspace{3pt} 
        {\includegraphics[width=\hsize,clip]{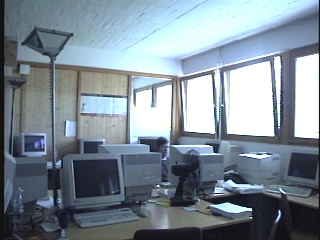}}
    \end{minipage}
    \vspace{3pt}
&\begin{minipage}{86pt}
/kyouyuuseki/\\(\textit{Shared desk})\\
\end{minipage}
&
1
&\begin{minipage}{80pt}
/nobasyanyanamae/\\/bawa\textbf{kyoo}/\\/\textbf{yuseki}/
\end{minipage}
&
1
&\begin{minipage}{80pt}
/\textbf{yuseki}nikibashita/\\/\textbf{kyoyuseki}/\\/\textbf{kyoyuseki}dayo/
\end{minipage}
&
1
&\begin{minipage}{70pt}
/kibashita/\\/ni/\\/\textbf{kyouyuseki}/
\end{minipage}
\\ \hline
    \begin{minipage}{58pt}
    \centering
    \vspace{3pt} 
        {\includegraphics[width=\hsize,clip]{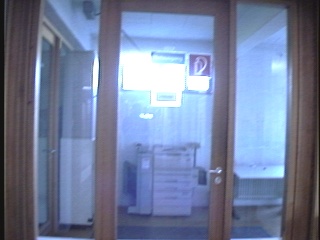}}
    \end{minipage}
    \vspace{3pt} 
&\begin{minipage}{86pt}
/ikidomari/\\(\textit{End of the corridor})\\
\end{minipage}
&
3
&\begin{minipage}{80pt}
/namaewa/\\/enikimashita/\\/gonobashiha/
\end{minipage}
&
3
&\begin{minipage}{80pt}
/dayo/\\/\textbf{ikidomari}/\\/fokoga/
\end{minipage}
&
3
&\begin{minipage}{70pt}
/\textbf{idomari}/\\/miriiNgusupesu/\\/koko/
\end{minipage}
\\ \hline
    \begin{minipage}{58pt}
    \centering
        \vspace{3pt} 
        {\includegraphics[width=\hsize,clip]{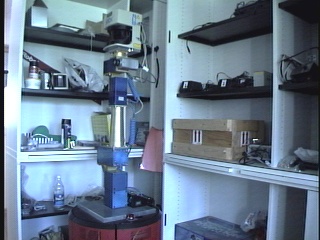}}
    \end{minipage}
        \vspace{3pt} 
&\begin{minipage}{86pt}
/roboqtookiba/\\(\textit{Robot storage space})\\
\end{minipage}
&
4
&\begin{minipage}{80pt}
/\textbf{robotokiba}/\\/\textbf{robotokiba}nya/\\/rimasu/
\end{minipage}
&
6
&\begin{minipage}{80pt}
/kochirawaaga\textbf{ro}/\\/\textbf{botoki}/\\/baninarimasu/
\end{minipage}
&
6
&\begin{minipage}{70pt}
/\textbf{wabotookiba}/\\/\textbf{robotookiba}/\\/kochiraga/
\end{minipage}
\\ \hline
\end{tabular}
    \caption{Top: Learning results of position distributions in a generated map. Ellipses denote the position distributions drawn on the map at steps 15, 30, and 50.
The colors of the ellipses were randomly determined for each index number $i_{t}=k$. Bottom: Examples of scene images captured by the robot. The correct word (in English) and estimated words are shown for each position distribution at steps 15, 30, and 50.} 
    \label{fig:map}
  \end{center}
\end{figure*}

\section{Experiment I}
\label{sec:exp}
We performed experiments to demonstrate online learning of spatial concepts in a novel environment.
In addition, we performed evaluations of place categorization and lexical acquisition related to places.
We compared the performance of the following methods: 
\begin{enumerate}
\item[(A)] SpCoSLAM \citep{ataniguchi_IROS2017} 
\item[(B)] SpCoSLAM with AW + WS (Section~\ref{sec:SpCoSLAM:learning}) 
\item[(C)] SpCoSLAM 2.0 (FLR--$i_{t},C_{t}$)
\item[(D)] SpCoSLAM 2.0 (FLR--$i_{t},C_{t}$ + RS)
\item[(E)] SpCoSLAM 2.0 (FLR--$i_{t},C_{t},S_{t}$ + SBU)
\item[(F)] SpCoA++ (Batch learning) \citep{taniguchi2018unsupervised}
\end{enumerate}

Methods (A) and (B) used the original and modified SpCoSLAM algorithms.
Methods (C) and (D) used the proposed improved algorithms under different conditions.
In methods (C) and (D), the lag value for FLR was set to $T_{L}=10$.
Method (E) used the proposed scalable algorithm under three different conditions: the lag values for the FLR were set to~{$T_{L}=$} 1, 10, and 20~{for (E1), (E2), and (E3), respectively}.
Batch-learning methods (F) was estimated by Gibbs sampling based on a weak-limit approximation \citep{fox2011sticky} of the Stick-Breaking Process (SBP) \citep{sethuraman1994constructive}, one of the constitutive methods of the Dirichlet Process (DP). 
The upper limits of the spatial concepts and position distributions were set to $L=50$ and $K=50$, respectively. %
We set the number of iterations for Gibbs sampling to $G=100$.
In method (F), we set the number of candidate word segmentation results for updating the language model to $M=6$, and the number of iterative estimation procedures to $I=10$.
In addition, (F) did not use image features in the same manner as the original model setting.
Note that SpCoA++ (F) was not evaluated in \citet{ataniguchi_IROS2017} because it is the latest batch-learning method.

\subsection{Online learning}
\label{sec:exp:online}
We conducted experiments of online spatial concept acquisition in a real environment.
We implemented SpCoSLAM 2.0 based on the open-source SpCoSLAM\footnote{\url{https://github.com/a-taniguchi/SpCoSLAM2}}, extending the gmapping package and implementing grid-based FastSLAM~2.0 \citep{gridbasedfastslam2007} in the Robot Operating System (ROS). 
We used an open dataset, albert-b-laser-vision, i.e., a rosbag file containing the odometry, laser range data, and image data.
This dataset was obtained from the Robotics Data Set Repository (Radish) \citep{Radish}. 
We prepared Japanese speech data corresponding to the movement of the robot from the above-mentioned dataset because speech data was not initially included.
The total number of taught utterances was $N=50$, including 10 types of phrases. 
The robot learned 10 places and 9 place names.
The microphone was a SHURE PG27-USB.
Julius dictation-kit-v4.4 (DNN-HMM decoding) \citep{lee2009recent} was used as a speech recognizer.
The initial word dictionary contained 115 Japanese syllables.
The unsupervised word segmentation system used latticelm \citep{neubig2012bayesian}.
The image feature extractor was implemented with Caffe, a deep-learning framework \citep{jia2014caffe}. 
We used a pre-trained CNN model, Places365-ResNet, trained with 365 scene categories from the Places2 Database with 1.8 million images \citep{zhou2017places}.
The number of particles was $R=30$.
The hyperparameters for online learning were set as follows: $\alpha=20$, $\gamma=0.1$, $\beta=0.1$, $\chi=0.1$, $m_{0}=[ 0 , 0 ]^{\rm T}$, $\kappa_{0}=0.001$, $V_{0}={\rm diag}(2,2)$, and $\nu_{0}=3$. 
The above-mentioned parameters were set such that all online methods were tested under the same conditions.
The hyperparameters for batch learning were set as follows: $\alpha=10$, $\gamma=10$, $\beta=0.1$, $m_{0}=[ 0 , 0 ]^{\rm T}$, $\kappa_{0}=0.001$, $V_{0}={\rm diag}(2,2)$, and $\nu_{0}=3$. 
The hyperparameters were determined manually and empirically according to each method.
Note that the speech recognition decoder, the image feature extractor, and the hyperparameters were changed from \citet{ataniguchi_IROS2017}.

Figure~\ref{fig:map} (top) shows the position distributions in the environmental maps at steps 15, 30, and 50 with (D).
This figure visualizes how spatial concepts are acquired during sequential mapping of the environment.
The position distributions were appropriately formed for places uttered by a user each time.
In step 15,~{the map covers} only 2 rooms (in the upper right) and a corridor, with 5 position distributions. 
The map obtained at step 50 covers the entire environment, and there were eventually 11 estimated position distributions.
Figure~\ref{fig:map} (bottom) shows an example of the correct phoneme sequence of the place name, and the three best words estimated by the probability distribution $p(S_{t} \mid i_{t}, \Theta_{t}, LM_{t})$ at step $t$.
The left side shows an example of the scene images observed in the $i_{t}$-th position distribution corresponding to the name of each place.
As the steps proceed, it can be seen that the words corresponding to the places were stably learned as phoneme sequences closer to the correct answers.
For example, in /kyouyuuseki/ (\textit{shared desk}), in step 15, the correspondence between the place and phoneme sequence was insufficiently learned: e.g., /bawa{kyoo}/ and /{yuseki}/. 
However, by step 50, the word was learned correctly: /{kyouyuseki}/.
The index of the position distribution of /roboqtookiba/ (\textit{robot storage space}) was changed from 4 to 6.
This change means that the label number switched as a result of the previous estimate values being modified while learning progressed.
Details of the online learning experiment can be found in a video online\footnote{\url{https://youtu.be/H5yztfmxGbc}}.

\subsection{Evaluation metrics}
\label{sec:exp:evaluation}
We evaluated the different algorithms according to the following metrics: the Adjusted Rand Index (ARI) \citep{hubert1985comparing} of the classification results of spatial concepts $C_{1:N}$ and position distribution $i_{1:N}$; the Estimation Accuracy Rate (EAR) of the estimated total numbers of spatial concepts $L$ and position distributions $K$; and the Phoneme Accuracy Rate (PAR) of uttered sentences and words related to places.
We conducted six learning trials under each algorithm condition.
The details of the evaluation metrics are described in the following sections.

\subsubsection{Estimation accuracy of spatial concepts} 
\label{sec:exp:evaluation:NMIEAR}
We compared the matching rate for the estimated indices $C_{1:N}$ of the spatial concept and the classification results of the correct answers given by a person.
In this experiment, the evaluation metric adopts the ARI, which is a measure of the similarity between two clustering results.
The matching rate for the estimated indices $i_{1:N}$ of the position distributions was evaluated in the same manner.

In addition, we evaluated the estimated number of spatial concepts $L$ and position distributions $K$ using the EAR.
The EAR was calculated as follows:
\begin{eqnarray}
{\rm EAR}= {\rm max} \left( 1 - \frac{\mid n^{\rm C}_{t} - n^{\rm E}_{t} \mid}{n^{\rm C}_{t}}, 0 \right)
 \label{eq:EAR}
\end{eqnarray}
where $n^{\rm C}_{t}$ is the correct number and $n^{\rm E}_{t}$ is the estimated number at time-step $t$.

\begin{table*}[!tb]
\begin{center}
\caption{Evaluation results {in a real environment.}}
\begin{tabular}{p{8pt}p{148pt}cccccccc} \hline
\multicolumn{2}{l}{Metric} & Improved & Scalable & \multicolumn{2}{c}{ARI} & \multicolumn{2}{c}{EAR} & \multicolumn{2}{c}{PAR} \\ 
\multicolumn{2}{l}{} &  &  & $C_{t}$ & $i_{t}$ & $L$ & $K$ & Sentence & Word \\ \hline 
(A)& SpCoSLAM &&& 
   {0.273} & {0.502} & {0.756} & {0.881} & {0.524} & {0.154} \\ 
({B})& SpCoSLAM with AW + WS &&& 
   {0.233} & {0.420} & {0.805} & {0.901} & {0.496} & {0.086} \\  
({C})& SpCoSLAM 2.0 (10 FLR--$i_{t},C_{t}$) &\checkmark&& 
   {0.324} & {0.602} & \underline{0.876} & \underline{0.913} & {0.533} & {0.157}  \\
({D})& SpCoSLAM 2.0 (10 FLR--$i_{t},C_{t}$ + RS)&\checkmark&& 
   {0.320} & {0.555} & \underline{\bf 0.881} & {0.901} & \underline{\bf 0.801} & \underline{0.419}  \\ 
({E1})& \begin{tabular}{l}SpCoSLAM 2.0 \\(1 FLR--$i_{t},C_{t},S_{t}$ + SBU)\end{tabular} &\checkmark&\checkmark& 
   {0.244} & {0.443} & {0.869} & \underline{\bf 0.923} & {0.648} & {0.158}  \\ 
({E2})& \begin{tabular}{l}SpCoSLAM 2.0 \\(10 FLR--$i_{t},C_{t},S_{t}$ + SBU)\end{tabular} &\checkmark&\checkmark& 
   {0.314} & {0.570} & {0.790} & {0.801} & {0.690} & {0.262}  \\ 
({E3})& \begin{tabular}{l}SpCoSLAM 2.0 \\(20 FLR--$i_{t},C_{t},S_{t}$ + SBU)\end{tabular} &\checkmark&\checkmark& 
   \underline{0.351} & \underline{\bf 0.673} & {0.748} & {0.890} & {0.704} & {0.292}  \\ \hline 
({F})& SpCoA++ (Batch learning) &&& 
   \underline{\bf 0.387} & \underline{0.624} & {0.700} & {0.648} & \underline{0.787} & \underline{\bf 0.524}  \\ \hline 
\end{tabular}
\label{table:hyouka2}
\end{center}
\end{table*}

\subsubsection{PAR of uttered sentences}
\label{sec:exp:evaluation:PARs}
We next compared the accuracy rate of phoneme recognition and word segmentation for all the recognized sentences.
However, it was difficult to separately weigh the ambiguous phoneme recognition and the unsupervised word segmentation. 
Therefore, the experiment considered the position of a delimiter as a single letter.
The correct phoneme sequence was suitably segmented into Japanese morphemes using MeCab \citep{kudo2006mecab}, an off-the-shelf Japanese morphological analyzer that is widely used for natural language processing. 
However, the name of the place was considered a single word.

We calculated the PAR of the uttered sentences with the correct phoneme sequence $s^{\rm P}_{t}$, and a phoneme sequence $s^{\rm R}_{t}$ of the recognition result of each uttered sentence.
The PAR was calculated as follows:
\begin{eqnarray}
{\rm PAR}= {\rm max} \left( 1 - \frac{{\rm LD}(s^{\rm P}_{t},s^{\rm R}_{t})}{n^{\rm p}}, 0 \right)
 \label{eq:PAR}
\end{eqnarray}
where ${\rm LD}()$ was calculated using the Levenshtein distance between $s^{\rm P}_{t}$ and $s^{\rm R}_{t}$. 
Here, $n^{\rm P}$ denotes the number of phonemes of the correct phoneme sequence.

\subsubsection{PAR of words related to places}
\label{sec:exp:evaluation:PARw}
We also evaluated whether a phoneme sequence has learned the properly segmented place names.
This experiment assumed a request for the best phoneme sequence, $s_{t}^{*}$, representing the self-position $x_{t}$ of the robot.
We compared the PAR of words with the correct place name and a selected word for each teaching place. 
The PAR was calculated using (\ref{eq:PAR}).

The selection of a word $s_{t,b}^{*}$ was calculated as follows:
\begin{eqnarray}
s_{t}^{*} = \argmax_{S_{t,b}} p(S_{t,b} \mid x_{t}, \Theta_{t}, LM_{t}).
 \label{eq:obest}
\end{eqnarray}
In this experiment, we used the self-position $x_{t}$ that was not included in the training data to evaluate the PAR of words. 
{Here, the robot can perform sufficiently accurate self-localization using a laser range finder.
Therefore, in this experiment, we assume that $x_{t}$ is given an accurate coordinate value without errors.}

The more a method accurately recognized words and acquired spatial concepts, the higher is the PAR.
We consider this evaluation metric to be an overall measure of the proposed method.

\begin{figure}[tb]
  \begin{center}
    \includegraphics[width=1.00\hsize]{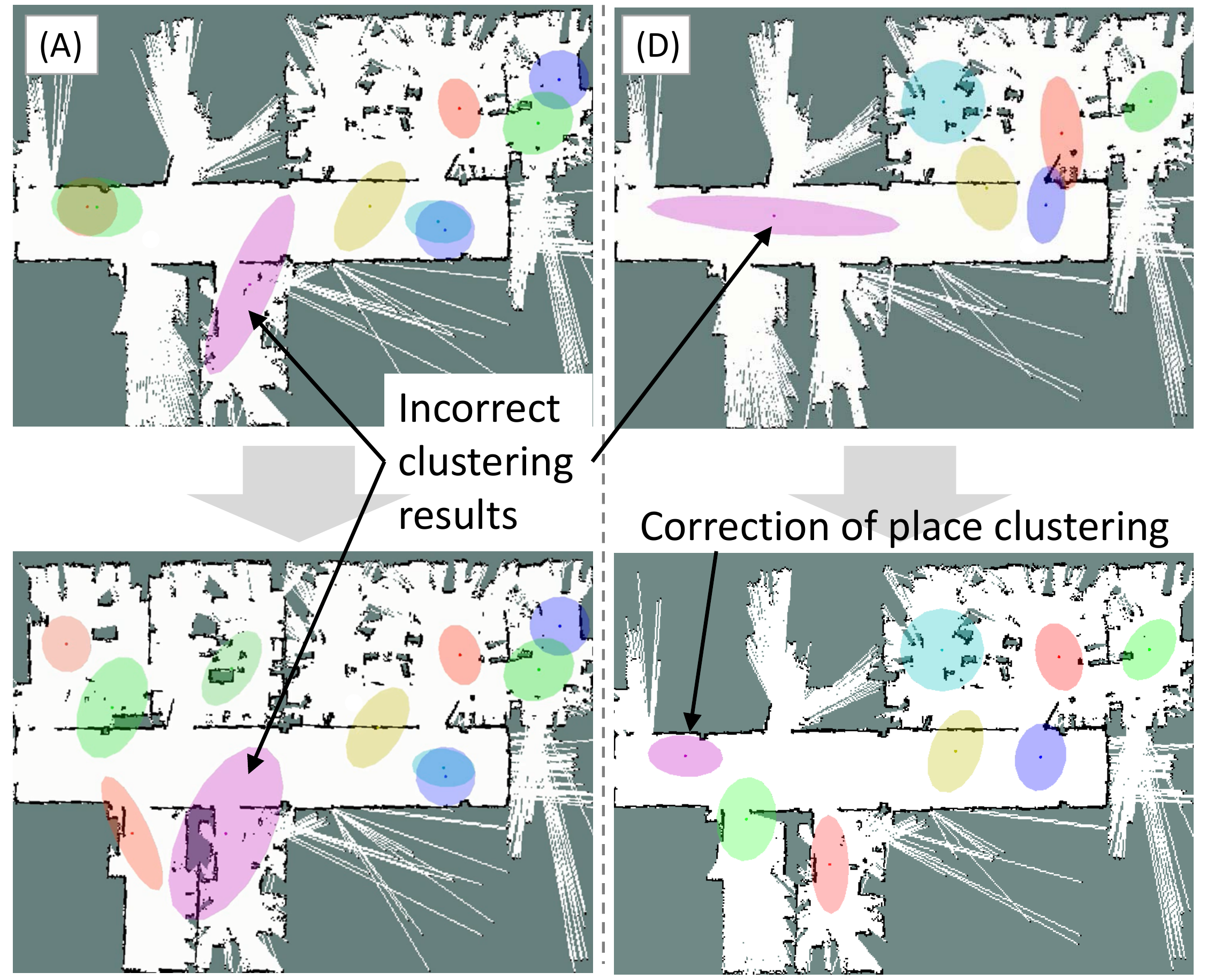}
    \caption{Examples of corrected place clustering results. Left: the original algorithm (A). Right: the improved SpCoSLAM 2.0 algorithm (D).}
    \label{fig:OriginalResult}
  \end{center}
\end{figure}

\begin{figure}[tb]
  \begin{center}
    \includegraphics[width=1.00\hsize]{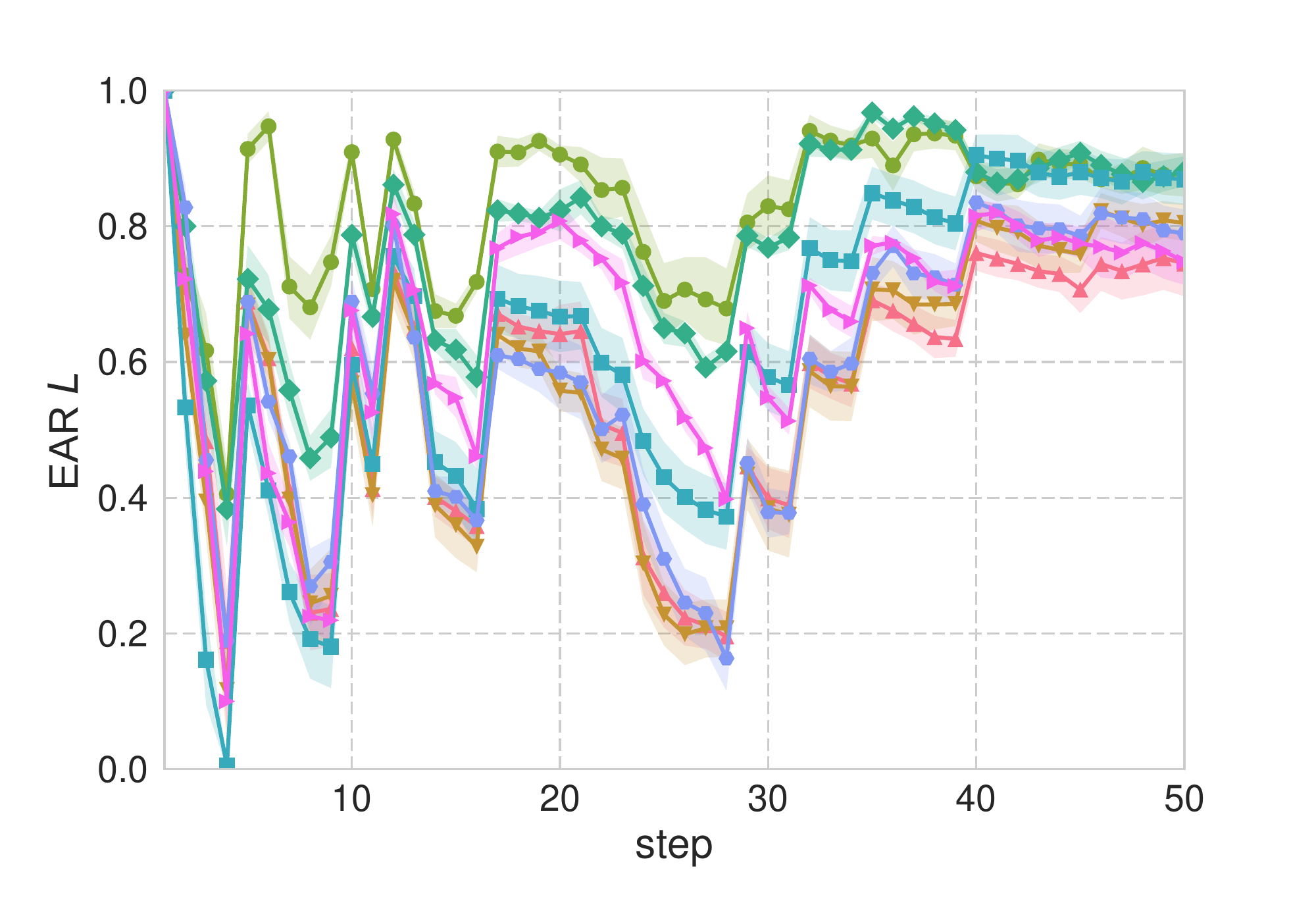}
    \\ \vspace{10pt}
    \includegraphics[width=1.00\hsize]{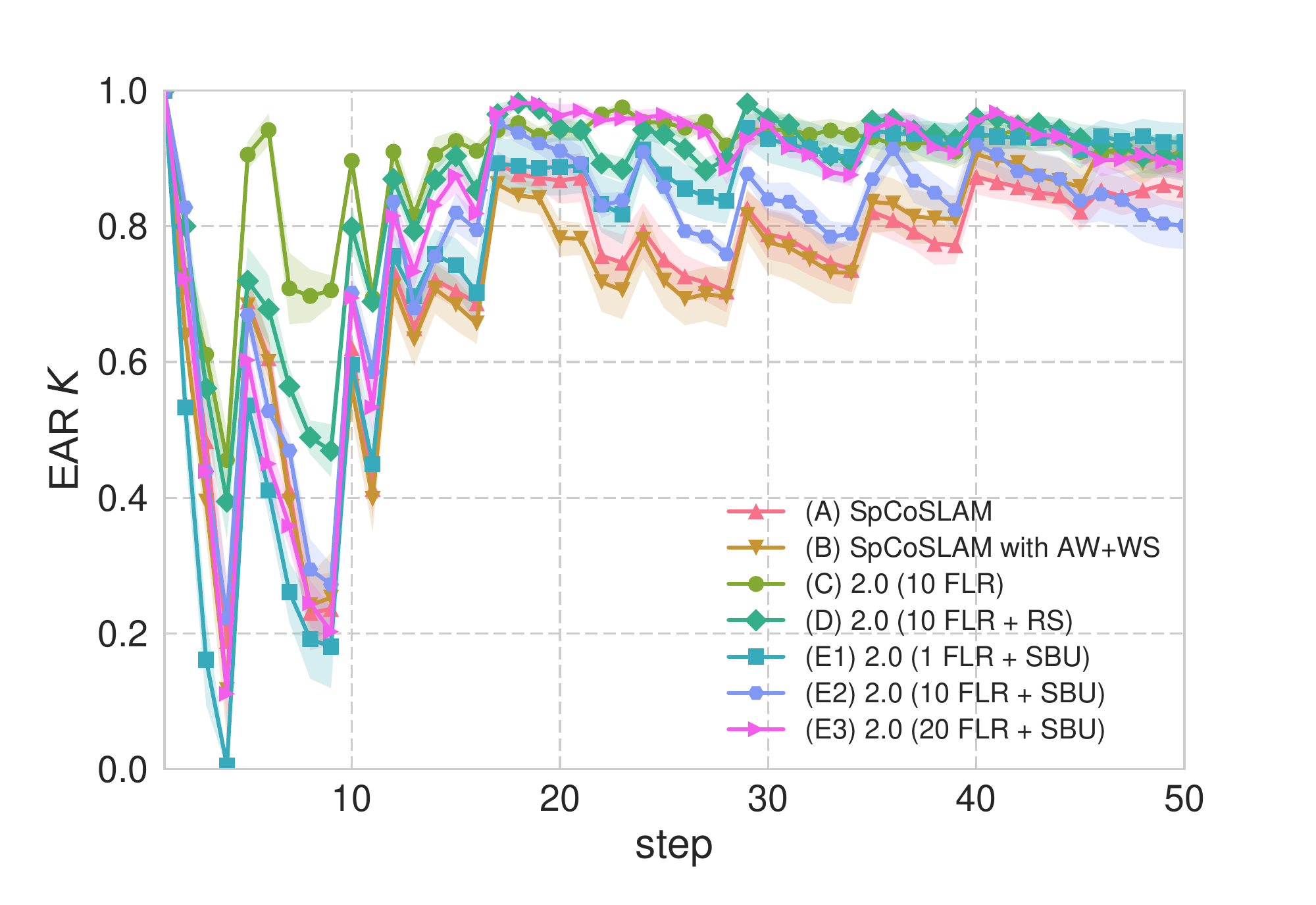}
    \caption{Change in the EAR regarding the estimated total number of spatial concepts $L$ (top) and position distributions $K$ (bottom) for each step.}
    \label{fig:EAR}
  \end{center}
\end{figure}

\begin{figure}[tb]
  \begin{center}
    \includegraphics[width=1.00\hsize]{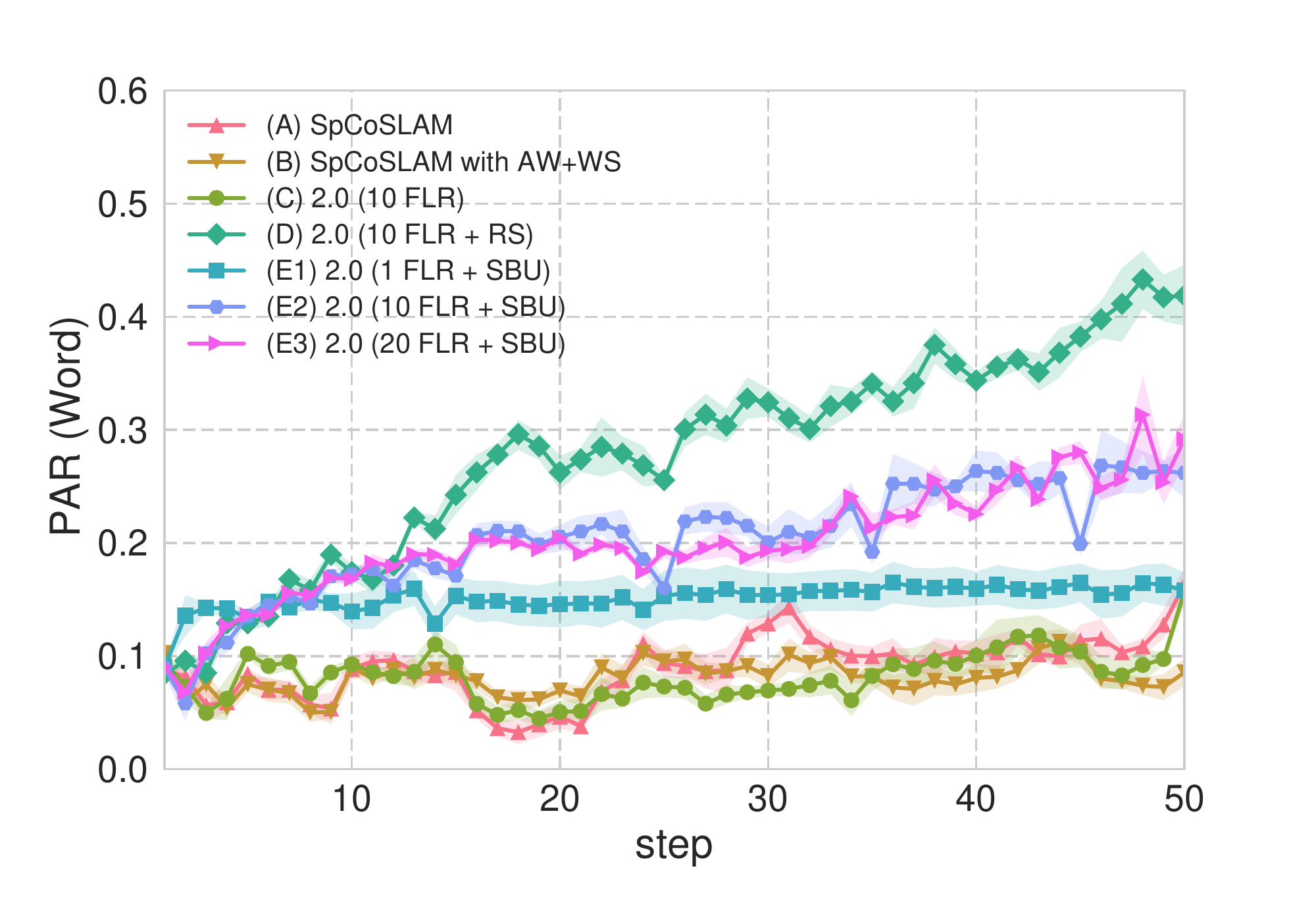}
    \caption{Change in the PAR of words for each step.}
    \label{fig:PARw}
  \end{center}
\end{figure}

\begin{table}[tb]
\begin{center}
\caption{Examples of word segmentation results of uttered sentences.}
\begin{tabular}{p{56pt}p{158pt}} \hline
English & 
{\it ``This place is {\bf the shared desk}.''} 
\\ 
Ground truth & 
kochira\,$\mid$\,ga\,$\mid$\,\textbf{kyouyuuseki}\,$\mid$\,ni\,$\mid$\,nari\,$\mid$\, masu  \\ \hline
(A) & 
a\,$\mid$\,kochiraga\textbf{gyoyusekiN}ni\,$\mid$\,narimasu \\ 
(D) & 
kochira\,$\mid$\,ga\,$\mid$\,\textbf{kyouyuseki}\,$\mid$\,ninarimasu  \\ 
(E3) & 
uo\,$\mid$\,kochi\,$\mid$\,ra\,$\mid$\,ga\,$\mid$\,\textbf{kyoyuseki}\,$\mid$\,nina\,$\mid$\,ri\,$\mid$\, ma\,$\mid$\,su   \\ 
(F) & 
ochiraga\,$\mid$\,\textbf{kyoyuseki}\,$\mid$\,ninarimasu \\ \hline
\multicolumn{2}{l}{\vspace{-2pt}}
\\ \hline
English & 
{\it ``This is {\bf the meeting space}.''} 
\\ 
Ground truth & 
koko\,$\mid$\,wa\,$\mid$\,\textbf{miitiNgusupeisu}\,$\mid$\,desu  \\ \hline
(A) &  
kokowaga\,$\mid$\,\textbf{midigisupesu}desujoouya  \\ 
(D) & 
kokowa\,$\mid$\,\textbf{miriiNgusupesu}\,$\mid$\,desu  \\ 
(E3) & 
kowa\,$\mid$\,\textbf{midigyusu}\,$\mid$\,\textbf{pesu}\,$\mid$\,desu  \\ 
(F) & 
gokoga\,$\mid$\,\textbf{miidiNgusupesu}\,$\mid$\,desu  \\ \hline
\multicolumn{2}{l}{\vspace{-2pt}}
\\ \hline
English &  
{\it ``{\bf The printer room} is here.''} 
\\
Ground truth & 
\textbf{puriNtaabeya}\,$\mid$\,wa\,$\mid$\,kochira\,$\mid$\,desu \\ \hline
(A) & 
io\textbf{poriNtabea}akochiragadesuduuryuzu qaqo \\ 
(D) & 
\textbf{puriNtabeya}\,$\mid$\,kochira\,$\mid$\,desu \\ 
(E3) & 
\textbf{puriNpabeya}\,$\mid$\,ta\,$\mid$\,kochiradesu \\ 
(F) & 
\textbf{poriNpabeya}\,$\mid$\,wakochiradesu \\ \hline
\end{tabular}
\label{bunkatu}
\end{center}
\end{table}

\subsection{Evaluation results and discussion}
\label{sec:exp:evaluation:result}
In this section, we discuss the improvement and scalability of the proposed learning algorithms.
Table~\ref{table:hyouka2} lists the averages of the evaluation values calculated using the metrics ARI, EAR, and PAR at step 50. 

\subsubsection{{ARI and EAR results}}
In terms of categorization accuracy, the proposed algorithms that introduced FLR tended to show higher ARI values than the original algorithms (A) and (B) of SpCoSLAM.
Figure~\ref{fig:OriginalResult} shows examples of the progress of place clustering for position distributions in (A) and (D).
The step numbers in the figures on the left (A) and right (D) are not the same. %
In these cases, large position distributions covering distant areas were learned, i.e., the purple ellipses in the figures on top.
In (A), incorrect clustering results were obtained during the final step (i.e., step 50) because the original SpCoSLAM algorithm cannot correct past erroneous estimations.
By contrast, in (D) by introducing FLR, an incorrect cluster occurred at step 25 (top right figure). 
However, the proposed algorithm could correct previous erroneous estimates at step 30 (bottom right figure).
Therefore, in the original algorithm (A), estimation errors adversely affect subsequent estimations. 
However, SpCoSLAM 2.0 (D) obtained more accurate estimations immediately, despite previous incorrect estimations.
{Similar situations to (D) were also confirmed in other proposed algorithms that introduced FLR.}
{Experimental results demonstrated  that FLR, which resamples the latent variables of the previous step using observations up to the current step, contributes to improving the accuracy of online place clustering.}

Figure~\ref{fig:EAR} shows the results of the EAR values with spatial concepts and position distributions, i.e., the accuracy of the estimated number of clusters, for each step.
The EAR values were not stable in the steps during the first half, although they converged stably to high values in the latter half.
In the result at step 50, (D) showed the highest EAR value $L$ and (E1) showed the highest EAR value $K$.
However, for both $L$ and $K$, looking at all the steps on average, (A) and (B) yielded relatively low values overall, and (C) and (D) yielded relatively high values.
(E1) -- (E3) tended to show values between original algorithms, (A) and (B), and improved algorithms with FLR, (C) and (D).
From the results of (C) and (D), EAR values improved considerably by introducing the FLR of $C_{t}$ and $i_{t}$.

\subsubsection{{PAR sentence and word results}}
From the results of the improved algorithm (D), the PAR values (sentence and word) improved markedly by adding the re-segmentation of the word sequences.
These results show that the robot can accurately segment the names of places and learn the relationship between places and words more precisely.
In particular, method (D), which combines the FLR and RS, achieved an overall improvement comparable to the other online algorithms.
Some trial results showed PAR values comparable to those of SpCoA++ (F).
Figure~\ref{fig:PARw} shows the PAR of words for each step.
The PAR tended to increase as a whole. 
Therefore, it can be expected that the PAR values will further increase as the number of steps advances.
Table~\ref{bunkatu} presents examples of word segmentation results with the four methods.
The correct phoneme sequence, i.e., ground truth, was segmented into Japanese morphemes using MeCab \citep{kudo2006mecab}, where
{``\,$\mid$\,''} denotes a delimiter, i.e., a word segment position.
The parts in bold correspond to the name for each place.
SpCoSLAM (A) showed under-segmentation results in many cases.
On the other hand, it can be seen that SpCoSLAM 2.0 (D) and (E3) properly segmented the phoneme sequences representing the name of the place.
Comparing (D) and (E3), (D) obtained segmentation results close to those of the batch learning method (F), and (E3) sometimes slightly over-segmented words.
Therefore, SpCoSLAM 2.0 can mitigate under-segmentation when the word segmentation of the batch learning method is applied in a pseudo-online manner.

\subsubsection{{Original and modified SpCoSLAM algorithms}}
Although the modified SpCoSLAM (B) is theoretically more appropriate than the original algorithm (A), few differences were found between them.
In the proposed algorithms, the time-driven process, i.e., SLAM part, and the event-driven process, i.e., spatial concept formation and lexical acquisition, were estimated by the same particle filter.
Although self-localization and mapping were performed each time the robot moved in an environment, latent variables for the spatial concepts and lexicon are updated only upon the user's utterance.
Thus, particles can fluctuate as a result of resampling due to movement in the absence of the user's utterance.
Consequently, the weight for self-localization might be influential, rather than the weight for the spatial concept and lexicon. 
This will be investigated in future work.

\begin{figure}[tb]
  \begin{center}
    \includegraphics[width=1.00\hsize]{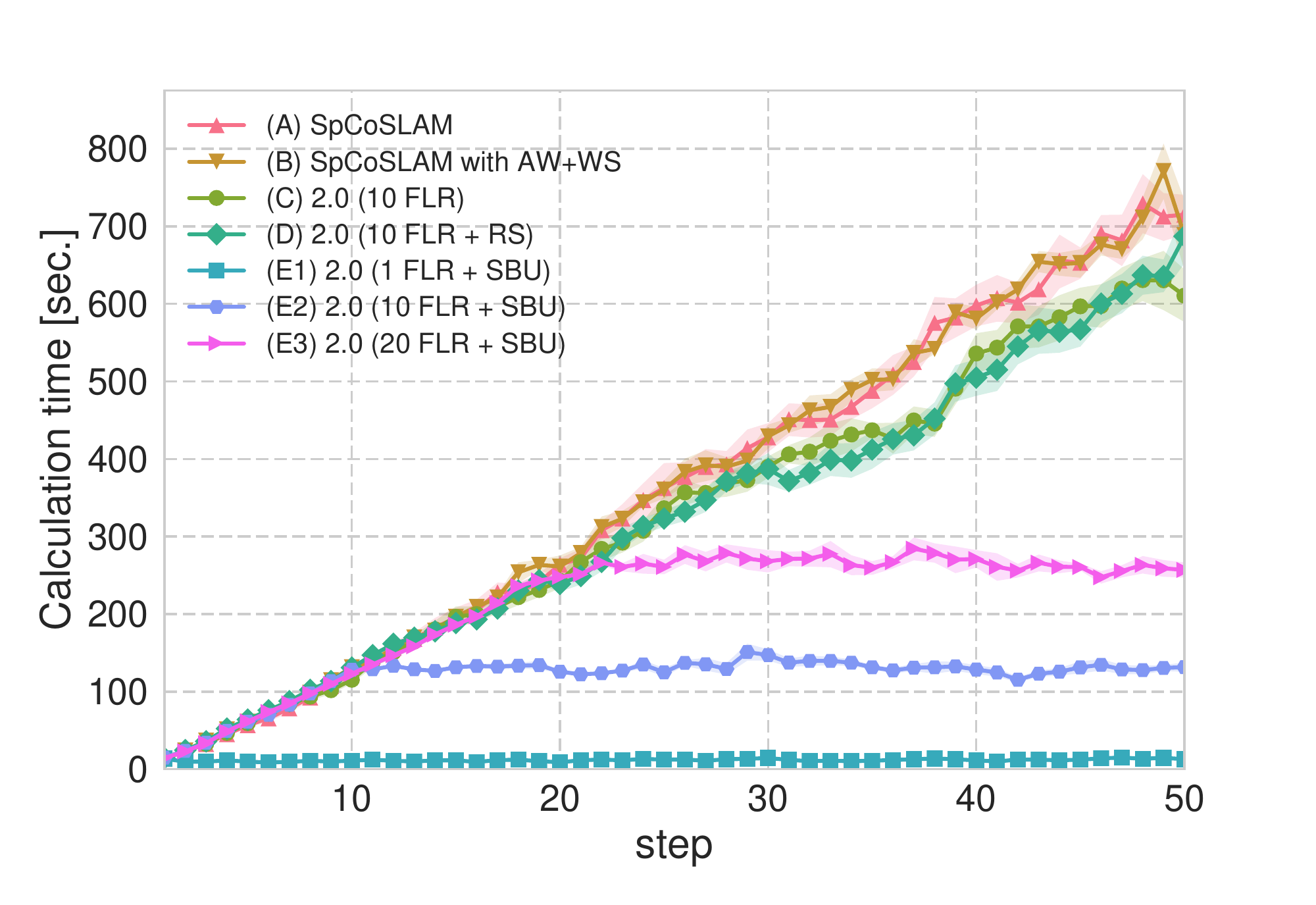}
    \caption{Calculation times par step for evaluating scalability.}
    \label{fig:time}
  \end{center}
\end{figure}

\subsubsection{{Calculation time and scalable algorithm}}
Figure~\ref{fig:time} shows the calculation times between online learning algorithms.
With batch learning, SpCoA++'s {overall calculation time including the runtime of rosbag for SLAM} was 13,850.873 s, and the calculation times per iteration for the iterative estimation procedure and Gibbs sampling were 1,318.954 s and 1.833 s, respectively. 
In the original SpCoSLAM algorithm, (A) and (B), and the improved SpCoSLAM 2.0 algorithm, (C) and (D), the calculation time increased with the number of steps, i.e., as the amount of data increased.
However, the scalable SpCoSLAM 2.0 algorithm (E1) -- (E3) retained a constant calculation time regardless of an increase in the amount of data.
Therefore, we can exert particularly powerful effects for long-term learning.

In the scalable algorithm (E1) -- (E3), the evaluation values of ARI and PAR tended to improve overall when the lag value increased. 
In particular, when the lag value was 20, relatively high evaluation values are seen to approach those of the improved algorithm.

Owing to a trade-off between the fixed-lag size and accuracy, the algorithm needs to be set appropriately according to both the computational power embedded in the robot and the duration requirements for actual operation.
In this experiment, we did not evaluate the scalability of the algorithm with parallel processing.
However, we considered that the proposed algorithm could be executed even faster by parallelizing the particle process and by using Graphics Processing Units (GPUs).
As such, {we consider that} the robot would be able to move within the environment while learning in real-time.

\section[Experiment II]{{Experiment II}} 
\label{sec:exp2}
In this experiment, it is investigated whether trends similar to the evaluation results of the real environmental dataset in Section~\ref{sec:exp} can be stably obtained across different environments.
Place categorization and lexical acquisition related to places in virtual home environments were evaluated, and the evaluation metrics ARI, EAR, and PAR for the methods (A) -- (F) were compared in the same manner as in Section~\ref{sec:exp}. 

\begin{figure}[tb]
  \begin{center}
    \includegraphics[width=0.480\hsize]{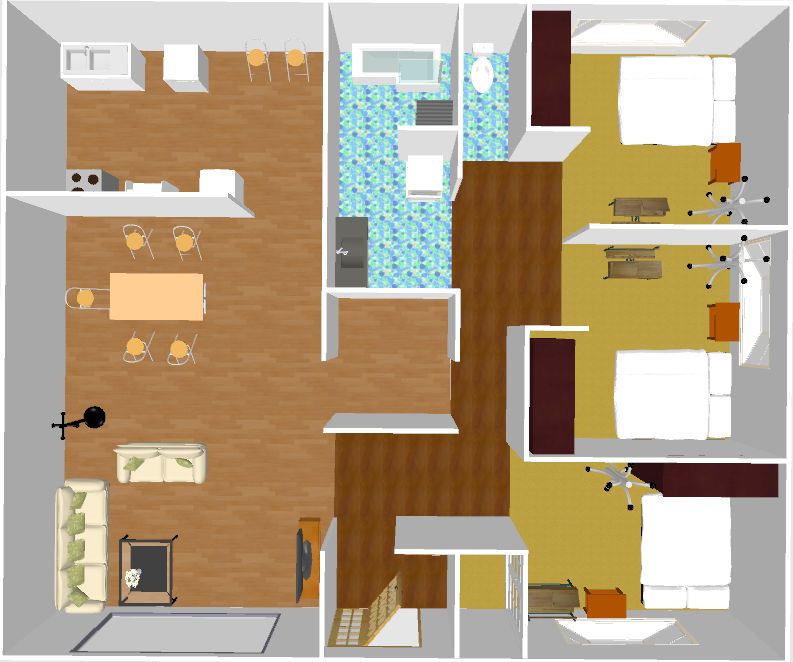}
    \hspace{6pt}
    \includegraphics[width=0.440\hsize]{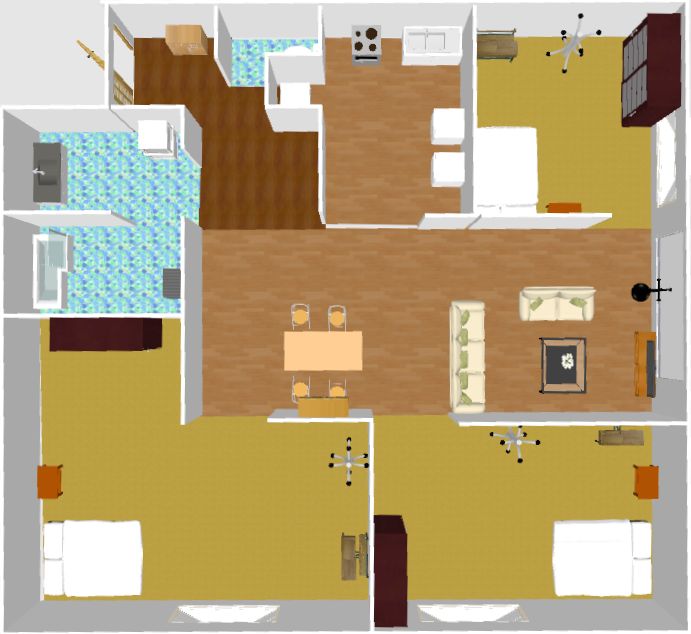}
    \caption{Examples of home environments in SIGVerse.}
    \label{fig:home}
  \end{center}
\end{figure}

\subsection{Condition}
\label{sec:exp2:condition}
Online spatial concept acquisition experiments were conducted in various virtual home environments.
The simulator environment was SIGVerse version 3.0~\citep{SIGVerse:SII2010}, a client-server based architecture that can connect the ROS and Unity.
The virtual robot in SIGVerse was Toyota's Human Support Robot (HSR), and we used 10 different home environments\footnote{3D models of home environments are available in \url{https://github.com/a-taniguchi/SweetHome3D_rooms}.} created using Sweet Home 3D\footnote{Sweet Home 3D: \url{http://www.sweethome3d.com/}}, which is a free software for interior design application.
Figure~\ref{fig:home} shows examples of the home environments.
For each place, 10 training data were provided on average.
The total number of taught utterances was $N = 60$, including 10 types of phrases. 
The robot learned six places and their respective names. 
The microphone and speech recognizer were the same as those in Section~\ref{sec:exp:online}.
The image feature extractor was a pre-trained BVLC CaffeNet model~\citep{jia2014caffe}.
The number of particles was $R = 10$.
The hyperparameters for learning were set as follows: $\alpha=10.0$, $\gamma=1.0$, $\beta=0.1$, $\chi=0.1$, $m_{0}=[ 0 , 0 ]^{\rm T}$, $\kappa_{0}=0.001$, $V_{0}={\rm diag}(2,2)$, and $\nu_{0}=3$. 
The hyperparameters were determined manually and empirically. 
The above-mentioned parameters were set such that all methods were tested under the same conditions.
In method (F), the upper limits of the spatial concepts and position distributions were set to $L = 20$ and $K = 20$, respectively. 
The other settings were identical to those in Section~\ref{sec:exp}.

The main target of the evaluation in this study is the accuracy of place clustering and lexical acquisition, i.e., extended points in SpCoSLAM 2.0.
Therefore, in this experiment, it is assumed that sufficiently accurate mapping and self-localization are possible with a high-precision distance sensor, and using an online learning algorithm which separates and omits the SLAM process was executed.
The true values obtained by the simulator were used as the self-position data.

\begin{table*}[!tb]
\begin{center}
\caption{Evaluation results in simulator environments}
\begin{tabular}{p{10pt}p{170pt}cccccccc} \hline
\multicolumn{2}{l}{Metric} & \multicolumn{2}{c}{ARI} & \multicolumn{2}{c}{EAR} & \multicolumn{2}{c}{PAR} \\ 
\multicolumn{2}{l}{} & $C_{t}$ & $i_{t}$ & $L$ & $K$ & Sentence & Word \\ \hline 
(A)& SpCoSLAM &
   {0.252} & {0.604} & {0.785} & {0.818} & {0.558} & {0.098} \\ 
({B})& SpCoSLAM with AW + WS &
   {0.347} & {0.684} & {0.802} & {0.815} & {0.565} & {0.141} \\  
({C})& SpCoSLAM 2.0 (10 FLR--$i_{t},C_{t}$) & 
   {0.346} & {0.713} & {0.733} & \underline{0.868} & {0.553} & {0.096} \\  
({D})& SpCoSLAM 2.0 (10 FLR--$i_{t},C_{t}$ + RS)&
   {0.314} & {0.719} & {0.730} & {0.840} & \underline{\bf 0.835} & \underline{0.464} \\  
({E1})& SpCoSLAM 2.0 (1 FLR--$i_{t},C_{t},S_{t}$ + SBU) &
   {0.307} & {0.672} & {0.817} & {0.800} & {0.671} & {0.165} \\  
({E2})& SpCoSLAM 2.0 (10 FLR--$i_{t},C_{t},S_{t}$ + SBU) &
   \underline{0.385} & {0.688} & \underline{0.833} & {0.782} & {0.733} & {0.305} \\  
({E3})& SpCoSLAM 2.0 (20 FLR--$i_{t},C_{t},S_{t}$ + SBU) &
   {0.354} & \underline{0.790} & \underline{\bf 0.883} & \underline{\bf 0.898} & {0.768} & {0.350} \\   \hline 
({F})& SpCoA++ (Batch learning) &
   \underline{\bf 0.522} & \underline{\bf 0.899} & {0.800} & {0.850} & \underline{0.830} & \underline{\bf 0.480} \\   \hline 
\end{tabular}
\label{table:hyouka2_sim}
\end{center}
\end{table*}

\begin{figure}[tb]
  \begin{center}
    \includegraphics[width=1.00\hsize]{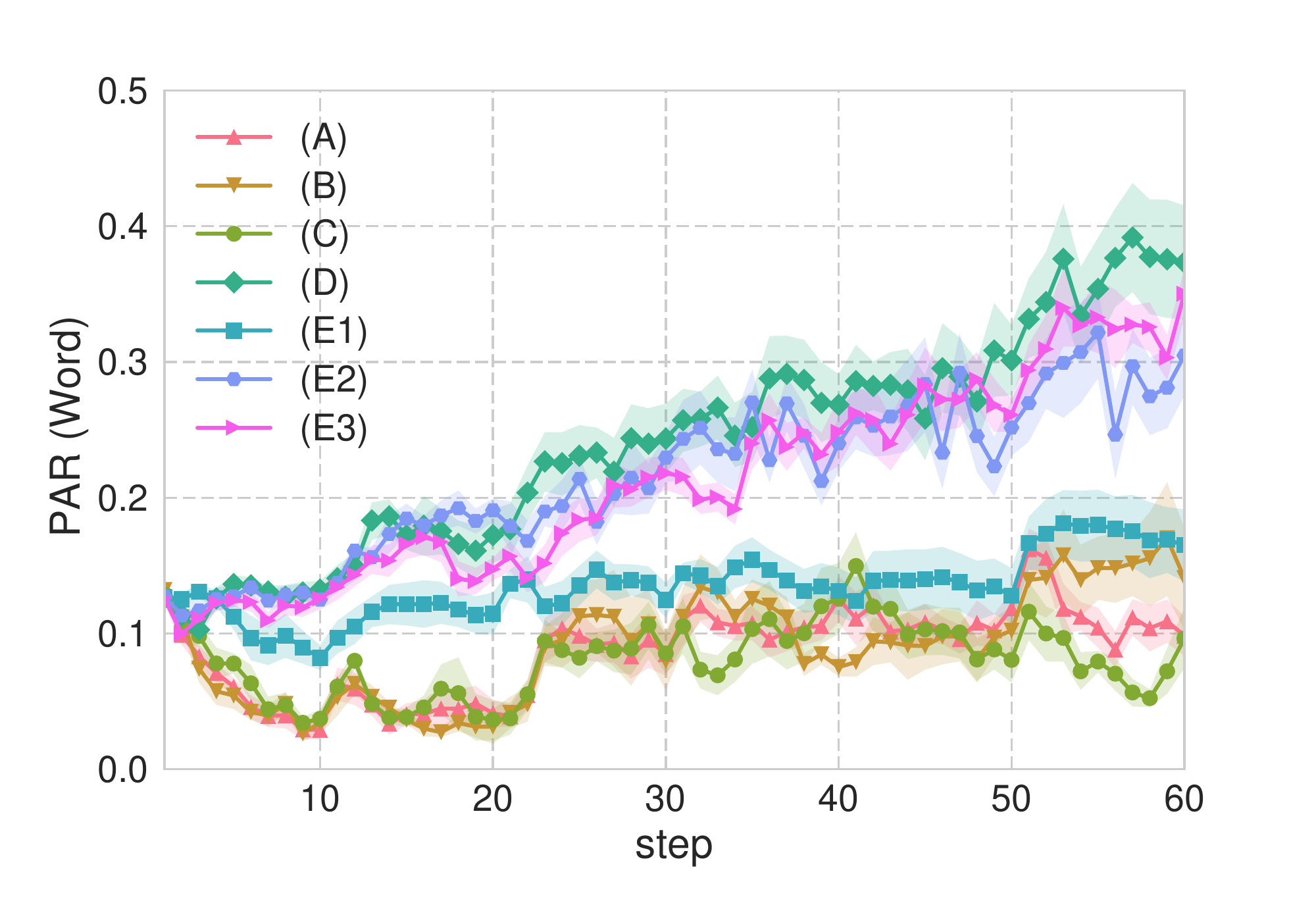}
    \caption{Change in PAR of words for each step in simulator environments.}
    \label{fig:PARw_SIGVerse}
  \end{center}
\end{figure}

\begin{figure}[tb]
  \begin{center}
    \includegraphics[width=1.00\hsize]{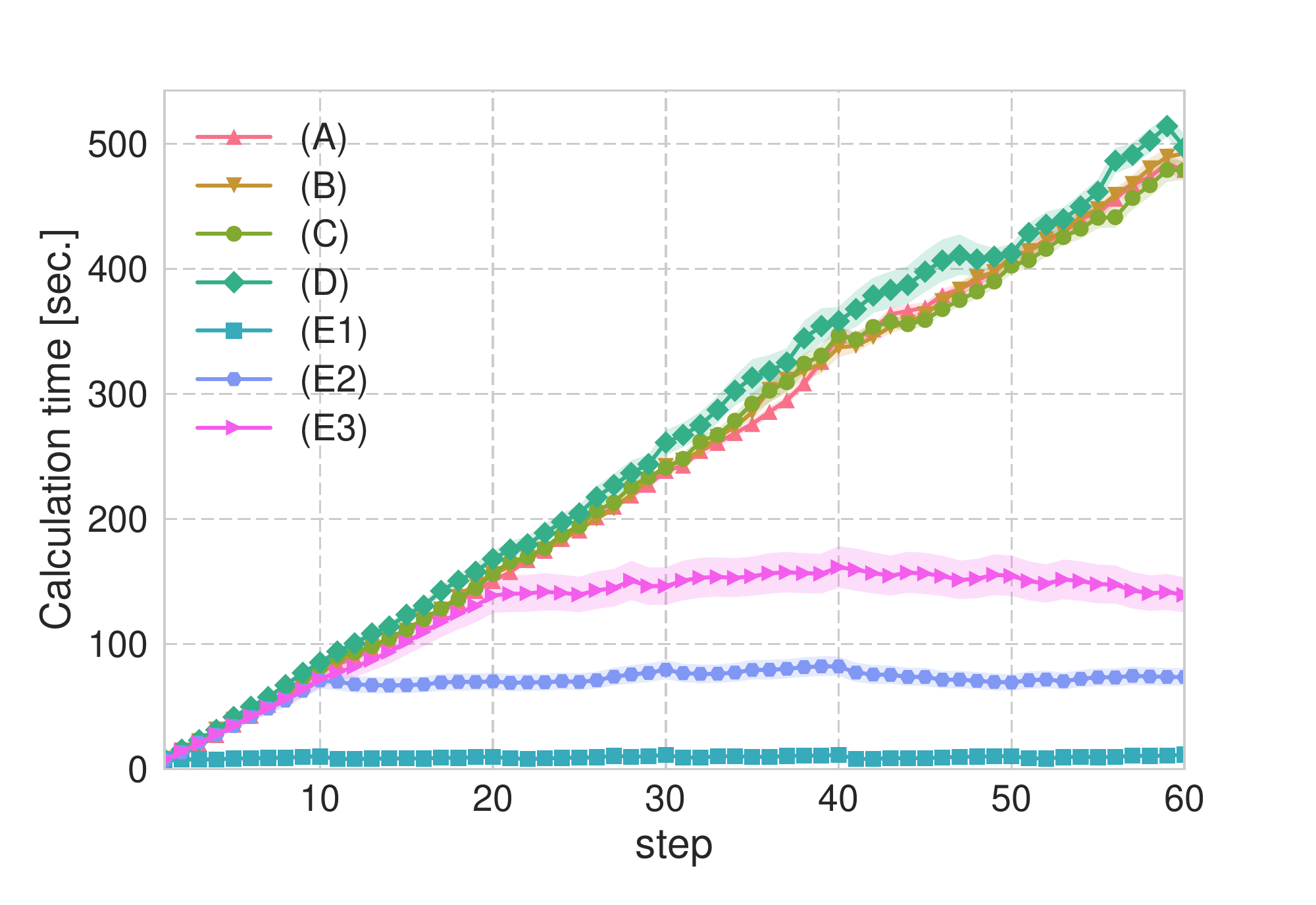}
    \caption{Calculation times par step in simulator environments.}
    \label{fig:time_SIGVerse}
  \end{center}
\end{figure}

\subsection{Result}
\label{sec:exp2:result}
In this section, the improvement and scalability of the proposed learning algorithms in home environments are discussed.
Table~\ref{table:hyouka2_sim} lists the averages of the evaluation values calculated using the metrics ARI, EAR, and PAR at step 60. 

The ARI showed a similar trend as the result of real environmental data.
However, compared to the original algorithms (A) and (B), there was almost no difference in the values in algorithms that introduced FLR.
In addition, the EAR showed a slightly different trend than the real environmental data.
In the improved algorithms (C) and (D), the number $L$ of categories of spatial concepts smaller than the true value was estimated compared to other algorithms.
We consider that this reason was due to the fact that it was re-combined into the same category by FLR.
Because the dataset was obtained in the simulator environment, for example, the image features could be insufficiently obtained for place categorization, i.e., similar features might be found in different places.
Such a problem did not occur when using real environmental data.

The PAR had the same tendency as the result of real environment data.
Similar to Section~\ref{sec:exp:evaluation:result}, the improved algorithm with RS (D) showed lexical acquisition accuracy comparable to batch learning (F).
In addition, the scalable algorithms with FLR of $S_{t}$ (E2) and (E3) showed higher values than the original algorithms.
Figure~\ref{fig:PARw_SIGVerse} shows the average values of the PAR of words for each step in different environments.
Similar to Fig.~\ref{fig:PARw}, the PAR tended to increase overall. 
Thus, it can be seen that RS and FLR of $S_{t}$ work effectively in virtual home environments.

In the comparison of the original and modified SpCoSLAM algorithms (A) and (B), the modified algorithm (B) showed higher overall values in the evaluation values of ARI and PAR.
We consider that the weight for the spatial concept and lexicon acted more directly in this experiment than in the experiment in Section~\ref{sec:exp}, because it was not affected by the weight for self-localization.

In scalable algorithms (E1) -- (E3), as the FLR value increased, the tendency for the overall evaluation values to increase appeared more prominently than for the results of real environment data.

Figure~\ref{fig:time_SIGVerse} shows the average calculation times between online learning algorithms in simulator environments.
We confirmed that the result was similar to Fig.~\ref{fig:time}, which was the result using the real environment data.
With batch learning, SpCoA++'s overall average calculation time was 8,076.288 s, and the calculation times per iteration for the iterative estimation procedure and Gibbs sampling were 807.623 s and 1.346 s, respectively. 

The following are the common inferences from the results of both the simulation and real-world environments.
For the online learning, if the user requires the performance of lexical acquisition even at an increased time cost, they can execute the improved algorithm (D) or scalable algorithm with a larger lag value, e.g., (E2) and (E3).
If the user requires high-speed calculation, they can obtain better results faster than the conventional algorithm (A) by executing a scalable algorithm such as (E1) and (E2).

\section{Conclusion}
\label{sec:conclusion}
This paper proposed an improved and scalable online learning algorithm to address the problems encountered by our previously proposed SpCoSLAM algorithm.
Specifically, we proposed online learning algorithm, called SpCoSLAM 2.0, for spatial concepts and lexical acquisition, for higher accuracy and scalability.
In experiments, we conducted online learning with a robot in a novel environment without any pre-existing lexicon and map.
In addition, we compared the proposed algorithm to the original online algorithm and to batch learning in terms of the estimation accuracy and calculation time.
The results demonstrate that the proposed algorithm is more accurate than the original algorithm and of comparable accuracy to batch learning.
Moreover, the calculation time of the proposed scalable algorithm becomes constant for each step, regardless of the amount of training data.
We expect this work to contribute to the realization of long-term spatial language interactions between humans and robots.

In the future, we shall experiment with long-term online learning of spatial concepts in large-scale environments based on the scalable algorithm proposed in this paper.
Furthermore, with additional development, it will be possible to introduce a forgetting mechanism to the proposed algorithm as with \citet{araki2012online}.
When a robot continues to operate over a long period of time it will encounter changes in the environment, such as the names of places and areas. Consequently, the robot will benefit from using the latest observation data as opposed to the previous observation data.
We believe that such a mechanism will be especially effective for long-term learning.

The proposed method constructs spatial concepts on a metric map; however, it can also be extended to learning the topological structure of places as with \citet{Karaoguz2016,Luperto2018}.
We explore whether this facilitates navigation tasks with human--robot linguistic interactions.
In addition, loop-closure detection has been studied actively in recent years, as is evident from long-term visual SLAM \citep{Han2018}.
The generative model of SpCoSLAM is connected to SLAM and lexical acquisition via latent variables related to the spatial concepts.
Therefore, we shall also explore loop-closure detection based on speech signals and investigate whether spatial concepts can positively affect mapping.

We will explore whether the SpCoSLAM model proposed herein can be integrated with other probabilistic models to form a large-scale cognitive model for general-purpose autonomous intelligent robots using a SERKET architecture \citep{nakamura2018serket}.
However, applications of the SERKET architecture are limited due to its computational cost for learning the enormous parameters of the whole model.
Even in such a case, we consider that our proposed approach to online learning will be extensively useful because it can be applied to various other Bayesian models.

\begin{acknowledgements}
{The authors thank Cyrill Stachniss for providing the dataset in the real robot.
The authors also thank Kazuya Asada and Keishiro Taguchi for providing virtual home environments and the dataset for spatial concepts in SIGVerse simulator.}

\end{acknowledgements}

\bibliographystyle{spbasic}      
\bibliography{./ForJabReff_utf8}

\begin{thebibliography}{50}
\providecommand{\natexlab}[1]{#1}
\providecommand{\url}[1]{{#1}}
\providecommand{\urlprefix}{URL }
\expandafter\ifx\csname urlstyle\endcsname\relax
  \providecommand{\doi}[1]{DOI~\discretionary{}{}{}#1}\else
  \providecommand{\doi}{DOI~\discretionary{}{}{}\begingroup
  \urlstyle{rm}\Url}\fi
\providecommand{\eprint}[2][]{\url{#2}}

\bibitem[{Aldous(1985)}]{aldous1985exchangeability}
Aldous D (1985) Exchangeability and related topics. {\'E}cole d'{\'E}t{\'e} de
  Probabilit{\'e}s de Saint{-}Flour XIII{-}1983 pp 1--198

\bibitem[{Aoki et~al.(2016)Aoki, Nishihara, Nakamura, and
  Nagai}]{aoki2016online}
Aoki T, Nishihara J, Nakamura T, Nagai T (2016) Online joint learning of object
  concepts and language model using multimodal hierarchical {D}irichlet
  process. In: Proceedings of the IEEE/RSJ International Conference on
  Intelligent Robots and Systems (IROS), IEEE, pp 2636--2642

\bibitem[{Araki et~al.(2012{\natexlab{a}})Araki, Nakamura, Nagai, Funakoshi,
  Nakano, and Iwahashi}]{araki2012online}
Araki T, Nakamura T, Nagai T, Funakoshi K, Nakano M, Iwahashi N
  (2012{\natexlab{a}}) Online object categorization using multimodal
  information autonomously acquired by a mobile robot. Advanced Robotics
  26(17):1995--2020

\bibitem[{Araki et~al.(2012{\natexlab{b}})Araki, Nakamura, Nagai, Nagasaka,
  Taniguchi, and Iwahashi}]{araki2012online_IROS}
Araki T, Nakamura T, Nagai T, Nagasaka S, Taniguchi T, Iwahashi N
  (2012{\natexlab{b}}) Online learning of concepts and words using multimodal
  {LDA} and hierarchical {P}itman-{Y}or {L}anguage {M}odel. In: Proceedings of
  the IEEE/RSJ International Conference on Intelligent Robots and Systems
  (IROS), IEEE, pp 1623--1630

\bibitem[{Ball et~al.(2013)Ball, Heath, Wiles, Wyeth, Corke, and
  Milford}]{ball2013openratslam}
Ball D, Heath S, Wiles J, Wyeth G, Corke P, Milford M (2013) Open{RatSLAM}: an
  open source brain-based slam system. Autonomous Robots 34(3):149--176

\bibitem[{Beevers and Huang(2007)}]{beevers2007fixed}
Beevers KR, Huang WH (2007) Fixed-lag sampling strategies for particle
  filtering slam. In: Proceedings of the IEEE International Conference on
  Robotics and Automation (ICRA), IEEE, pp 2433--2438

\bibitem[{B{\"o}rschinger and Johnson(2011)}]{borschinger2011particle}
B{\"o}rschinger B, Johnson M (2011) A particle filter algorithm for {B}ayesian
  wordsegmentation. In: Australasian Language Technology Association Workshop
  2011, Citeseer, p~10

\bibitem[{B{\"o}rschinger and Johnson(2012)}]{borschinger2012using}
B{\"o}rschinger B, Johnson M (2012) Using rejuvenation to improve particle
  filtering for {B}ayesian word segmentation. In: Proceedings of the 50th
  Annual Meeting of the Association for Computational Linguistics, Association
  for Computational Linguistics, pp 85--89

\bibitem[{Cangelosi and Schlesinger(2015)}]{cangelosi2015developmental}
Cangelosi A, Schlesinger M (2015) Developmental Robotics: From Babies to
  Robots. Intelligent Robotics and Autonomous Agents Series, MIT Press,
  \urlprefix\url{https://books.google.co.jp/books?id=AbKPoAEACAAJ}

\bibitem[{Canini et~al.(2009)Canini, Shi, and Griffiths}]{canini2009online}
Canini KR, Shi L, Griffiths TL (2009) Online inference of topics with latent
  {D}irichlet allocation. In: Proceedings of the International Conference on
  Artificial Intelligence and Statistics (AISTATS), vol~9, pp 65--72

\bibitem[{Doucet et~al.(2000)Doucet, De~Freitas, Murphy, and
  Russell}]{doucet2000rao}
Doucet A, De~Freitas N, Murphy K, Russell S (2000) {R}ao-{B}lackwellised
  particle filtering for dynamic {B}ayesian networks. In: Proceedings of the
  16th Conference on Uncertainty in artificial intelligence, Morgan Kaufmann
  Publishers Inc., pp 176--183

\bibitem[{Fox et~al.(2011)Fox, Sudderth, Jordan, and Willsky}]{fox2011sticky}
Fox EB, Sudderth EB, Jordan MI, Willsky AS (2011) A sticky {HDP-HMM} with
  application to speaker diarization. The Annals of Applied Statistics
  5(2A):1020--1056

\bibitem[{Grisetti et~al.(2007)Grisetti, Stachniss, and
  Burgard}]{gridbasedfastslam2007}
Grisetti G, Stachniss C, Burgard W (2007) Improved techniques for grid mapping
  with {R}ao-{B}lackwellized particle filters. IEEE Transactions on Robotics
  23:34--46

\bibitem[{Gu et~al.(2016)Gu, Taguchi, Hattori, Hoguro, and
  Umezaki}]{gu2016learning}
Gu Z, Taguchi R, Hattori K, Hoguro M, Umezaki T (2016) Learning of relative
  spatial concepts from ambiguous instructions. In: Proceedings of the 13th
  IFAC/IFIP/IFORS/IEA Symposium on Analysis, Design, and Evaluation of
  Human-Machine Systems (IFAC HMS), Elsevier, vol~49, pp 150--153

\bibitem[{Hagiwara et~al.(2018)Hagiwara, Inoue, Kobayashi, and
  Taniguchi}]{hagiwara2018hierarchical}
Hagiwara Y, Inoue M, Kobayashi H, Taniguchi T (2018) Hierarchical spatial
  concept formation based on multimodal information for human support robots.
  Frontiers in Neurorobotics 12:11, \doi{10.3389/fnbot.2018.00011}

\bibitem[{Han et~al.(2018)Han, Wang, Huang, and Zhang}]{Han2018}
Han F, Wang H, Huang G, Zhang H (2018) Sequence-based sparse optimization
  methods for long-term loop closure detection in visual slam. Autonomous
  Robots 42(7):1323--1335, \doi{10.1007/s10514-018-9736-3}

\bibitem[{Heath et~al.(2016)Heath, Ball, and Wiles}]{heathlingodroids2016}
Heath S, Ball D, Wiles J (2016) Lingodroids: Cross-situational learning for
  episodic elements. IEEE Transactions on Cognitive and Developmental Systems
  8(1):3--14, \doi{10.1109/TAMD.2015.2442619}

\bibitem[{Hemachandra et~al.(2014)Hemachandra, Walter, Tellex, and
  Teller}]{hemachandra2014learning}
Hemachandra S, Walter MR, Tellex S, Teller S (2014) Learning spatial-semantic
  representations from natural language descriptions and scene classifications.
  In: Proceedings of the IEEE International Conference on Robotics and
  Automation (ICRA), IEEE, pp 2623--2630

\bibitem[{Howard and Roy(2003)}]{Radish}
Howard A, Roy N (2003) The robotics data set repository (radish).
  \urlprefix\url{http://radish.sourceforge.net/}

\bibitem[{Hubert and Arabie(1985)}]{hubert1985comparing}
Hubert L, Arabie P (1985) Comparing partitions. Journal of classification
  2(1):193--218

\bibitem[{Inamura et~al.(2010)Inamura, Shibata, Sena, Hashimoto, Kawai,
  Miyashita, Sakurai, Shimizu, Otake, Hosoda et~al.}]{SIGVerse:SII2010}
Inamura T, Shibata T, Sena H, Hashimoto T, Kawai N, Miyashita T, Sakurai Y,
  Shimizu M, Otake M, Hosoda K, et~al. (2010) Simulator platform that enables
  social interaction simulation --{SIGV}erse: {S}ocio{I}ntelli{G}enesis
  simulator--. In: Proceedings of the IEEE/SICE International Symposium on
  System Integration, pp 212--217

\bibitem[{Isobe et~al.(2017)Isobe, Taniguchi, Hagiwara, and
  Taniguchi}]{isobe2017learning}
Isobe S, Taniguchi A, Hagiwara Y, Taniguchi T (2017) Learning relationships
  between objects and places by multimodal spatial concept with bag of objects.
  In: Proceedings of the International Conference on Social Robotics (ICSR),
  Springer, pp 115--125

\bibitem[{Jia et~al.(2014)Jia, Shelhamer, Donahue, Karayev, Long, Girshick,
  Guadarrama, and Darrell}]{jia2014caffe}
Jia Y, Shelhamer E, Donahue J, Karayev S, Long J, Girshick R, Guadarrama S,
  Darrell T (2014) Caffe: Convolutional architecture for fast feature
  embedding. arXiv preprint arXiv:14085093

\bibitem[{Kantas et~al.(2015)Kantas, Doucet, Singh, Maciejowski, Chopin
  et~al.}]{kantas2015particle}
Kantas N, Doucet A, Singh SS, Maciejowski J, Chopin N, et~al. (2015) On
  particle methods for parameter estimation in state-space models. Statistical
  science 30(3):328--351

\bibitem[{Karao{\u{g}}uz and Bozma(2016)}]{Karaoguz2016}
Karao{\u{g}}uz H, Bozma HI (2016) An integrated model of autonomous topological
  spatial cognition. Autonomous Robots 40(8):1379--1402,
  \doi{10.1007/s10514-015-9514-4}

\bibitem[{Kitagawa(2014)}]{kitagawa2014computational}
Kitagawa G (2014) Computational aspects of sequential {M}onte {C}arlo filter
  and smoother. Annals of the Institute of Statistical Mathematics
  66(3):443--471

\bibitem[{Kostavelis and Gasteratos(2015)}]{kostavelis2015semantic}
Kostavelis I, Gasteratos A (2015) Semantic mapping for mobile robotics tasks: A
  survey. Robotics and Autonomous Systems 66:86--103

\bibitem[{Krizhevsky et~al.(2012)Krizhevsky, Sutskever, and
  Hinton}]{krizhevsky2012imagenet}
Krizhevsky A, Sutskever I, Hinton G (2012) Imagenet classification with deep
  convolutional neural networks. In: Proceedings of the Advances in Neural
  Information Processing Systems (NIPS), Nevada, United States, pp 1097--1105

\bibitem[{Kudo(2006)}]{kudo2006mecab}
Kudo T (2006) Me{C}ab: Yet another part-of-speech and morphological analyzer.
  \urlprefix\url{https://github.com/taku910/mecab}

\bibitem[{Landsiedel et~al.(2017)Landsiedel, Rieser, Walter, and
  Wollherr}]{landsiedel2017review}
Landsiedel C, Rieser V, Walter M, Wollherr D (2017) A review of spatial
  reasoning and interaction for real-world robotics. Advanced Robotics
  31(5):222--242

\bibitem[{Lee and Kawahara(2009)}]{lee2009recent}
Lee A, Kawahara T (2009) Recent development of open-source speech recognition
  engine {J}ulius. In: Proceedings of the APSIPA ASC, pp 131--137

\bibitem[{Luperto and Amigoni(2018)}]{Luperto2018}
Luperto M, Amigoni F (2018) Predicting the global structure of indoor
  environments: A constructive machine learning approach. Autonomous Robots
  \doi{10.1007/s10514-018-9732-7}

\bibitem[{Mochihashi et~al.(2009)Mochihashi, Yamada, and
  Ueda}]{mochihashi2009bayesian}
Mochihashi D, Yamada T, Ueda N (2009) {B}ayesian unsupervised word segmentation
  with nested {Pitman-Yor} language modeling. In: Proceedings of the Joint
  Conference of the 47th Annual Meeting of the ACL and the 4th International
  Joint Conference on Natural Language Processing of the AFNLP (ACL-IJCNLP), pp
  100--108

\bibitem[{Montemerlo et~al.(2003)Montemerlo, Thrun, Koller, Wegbreit
  et~al.}]{montemerlo2003fastslam}
Montemerlo M, Thrun S, Koller D, Wegbreit B, et~al. (2003) {FastSLAM} 2.0: An
  improved particle filtering algorithm for simultaneous localization and
  mapping that provably converges. In: Proceedings of the International Joint
  Conference on Artificial Intelligence (IJCAI), pp 1151--1156

\bibitem[{Nakamura et~al.(2018)Nakamura, Nagai, and
  Taniguchi}]{nakamura2018serket}
Nakamura T, Nagai T, Taniguchi T (2018) Serket: An architecture for connecting
  stochastic models to realize a large-scale cognitive model. Frontiers in
  Neurorobotics 12:25, \doi{10.3389/fnbot.2018.00025}

\bibitem[{Neubig et~al.(2012)Neubig, Mimura, and Kawahara}]{neubig2012bayesian}
Neubig G, Mimura M, Kawahara T (2012) {B}ayesian learning of a language model
  from continuous speech. IEICE Transactions on Information and Systems
  95(2):614--625

\bibitem[{Nishihara et~al.(2017)Nishihara, Nakamura, and
  Nagai}]{nishihara2017online}
Nishihara J, Nakamura T, Nagai T (2017) Online algorithm for robots to learn
  object concepts and language model. IEEE Transactions on Cognitive and
  Developmental Systems 9(3):255--268, \doi{10.1109/TCDS.2016.2552579}

\bibitem[{Pronobis and Jensfelt(2012)}]{pronobis2012large}
Pronobis A, Jensfelt P (2012) Large-scale semantic mapping and reasoning with
  heterogeneous modalities. In: Proceedings of the IEEE International
  Conference on Robotics and Automation (ICRA), IEEE, pp 3515--3522

\bibitem[{Rangel et~al.(2018)Rangel, Cazorla, Garc{\'i}a-Varea,
  Romero-Gonz{\'a}lez, and Mart{\'i}nez-G{\'o}mez}]{Rangel2018}
Rangel JC, Cazorla M, Garc{\'i}a-Varea I, Romero-Gonz{\'a}lez C,
  Mart{\'i}nez-G{\'o}mez J (2018) Automatic semantic maps generation from
  lexical annotations. Autonomous Robots \doi{10.1007/s10514-018-9723-8}

\bibitem[{Sethuraman(1994)}]{sethuraman1994constructive}
Sethuraman J (1994) A constructive definition of {D}irichlet priors. Statistica
  Sinica 4:639--650

\bibitem[{S{\"u}nderhauf et~al.(2016)S{\"u}nderhauf, Dayoub, McMahon, Talbot,
  Schulz, Corke, Wyeth, Upcroft, and Milford}]{sunderhauf2016place}
S{\"u}nderhauf N, Dayoub F, McMahon S, Talbot B, Schulz R, Corke P, Wyeth G,
  Upcroft B, Milford M (2016) Place categorization and semantic mapping on a
  mobile robot. In: Proceedings of the IEEE International Conference on
  Robotics and Automation (ICRA), IEEE, pp 5729--5736

\bibitem[{Taguchi et~al.(2011)Taguchi, Yamada, Hattori, Umezaki, Hoguro,
  Iwahashi, Funakoshi, and Nakano}]{taguchi2011learning}
Taguchi R, Yamada Y, Hattori K, Umezaki T, Hoguro M, Iwahashi N, Funakoshi K,
  Nakano M (2011) Learning place-names from spoken utterances and localization
  results by mobile robot. In: Proceedings of the Annual Conference of the
  International Speech Communication Association (INTERSPEECH), pp 1325--1328

\bibitem[{Taniguchi et~al.(2016)Taniguchi, Taniguchi, and
  Inamura}]{taniguchi_spcoa}
Taniguchi A, Taniguchi T, Inamura T (2016) Spatial concept acquisition for a
  mobile robot that integrates self{-}localization and unsupervised word
  discovery from spoken sentences. IEEE Transactions on Cognitive and
  Developmental Systems 8(4):285--297, \doi{10.1109/TCDS.2016.2565542}

\bibitem[{Taniguchi et~al.(2017)Taniguchi, Hagiwara, Taniguchi, and
  Inamura}]{ataniguchi_IROS2017}
Taniguchi A, Hagiwara Y, Taniguchi T, Inamura T (2017) Online spatial concept
  and lexical acquisition with simultaneous localization and mapping. In:
  Proceedings of the IEEE/RSJ International Conference on Intelligent Robots
  and Systems (IROS), pp 811--818, \doi{10.1109/IROS.2017.8202243}

\bibitem[{Taniguchi et~al.(2018{\natexlab{a}})Taniguchi, Taniguchi, and
  Inamura}]{taniguchi2018unsupervised}
Taniguchi A, Taniguchi T, Inamura T (2018{\natexlab{a}}) Unsupervised spatial
  lexical acquisition by updating a language model with place clues. Robotics
  and Autonomous Systems 99:166--180, \doi{10.1016/j.robot.2017.10.013}

\bibitem[{Taniguchi et~al.(2018{\natexlab{b}})Taniguchi, Ugur, Hoffmann,
  Jamone, Nagai, Rosman, Matsuka, Iwahashi, Oztop, Piater, and
  Wörgötter}]{taniguchi2018TCDSsurvey}
Taniguchi T, Ugur E, Hoffmann M, Jamone L, Nagai T, Rosman B, Matsuka T,
  Iwahashi N, Oztop E, Piater J, Wörgötter F (2018{\natexlab{b}}) Symbol
  emergence in cognitive developmental systems: a survey. IEEE Transactions on
  Cognitive and Developmental Systems pp 1--1, \doi{10.1109/TCDS.2018.2867772}

\bibitem[{Thrun et~al.(2005)Thrun, Burgard, and Fox}]{thrun2005probabilistic}
Thrun S, Burgard W, Fox D (2005) Probabilistic Robotics. MIT Press

\bibitem[{Ueda et~al.(2016)Ueda, Mizuta, Yamakawa, and
  Okada}]{ueda2016particle}
Ueda R, Mizuta K, Yamakawa H, Okada H (2016) Particle filter on episode for
  learning decision making rule. In: Proceedings of the International
  Conference on Intelligent Autonomous Systems (IAS), Springer, pp 737--754

\bibitem[{Walter et~al.(2013)Walter, Hemachandra, Homberg, Tellex, and
  Teller}]{walter2013learning}
Walter MR, Hemachandra S, Homberg B, Tellex S, Teller S (2013) Learning
  semantic maps from natural language descriptions. In: Proceedings of
  Robotics: Science and Systems (RSS)

\bibitem[{Zhou et~al.(2018)Zhou, Lapedriza, Khosla, Oliva, and
  Torralba}]{zhou2017places}
Zhou B, Lapedriza A, Khosla A, Oliva A, Torralba A (2018) Places: A 10 million
  image database for scene recognition. IEEE Transactions on Pattern Analysis
  and Machine Intelligence 40(6):1452--1464

\end{thebibliography}
\end{document}